\documentclass[lettersize,journal]{IEEEtran}
\usepackage{amsmath,amsfonts}
\usepackage{algorithmic}
\usepackage{algorithm}
\usepackage{array}
\usepackage[caption=false,font=normalsize,labelfont=sf,textfont=sf]{subfig}
\usepackage{textcomp}
\usepackage{stfloats}
\usepackage{url}
\usepackage{verbatim}
\usepackage{graphicx}
\usepackage{cite}
\usepackage{enumitem}
\usepackage{listings}
\usepackage{xcolor}
\usepackage{multirow}
\usepackage{graphicx}
\usepackage{color}
\definecolor{codegray}{RGB}{150,150,150}
\definecolor{codebg}{RGB}{245,245,245}
\definecolor{codeframe}{RGB}{220,220,220}
\definecolor{codeblue}{RGB}{0,70,140}
\definecolor{codered}{RGB}{160,0,0}

\lstdefinestyle{customstyle}{
  basicstyle=\ttfamily\footnotesize,  
  backgroundcolor=\color{codebg},     
  frame=single,                       
  rulecolor=\color{codeframe},
  framerule=0.4pt,
  columns=fullflexible,
  keepspaces=true,
  showstringspaces=false,
  tabsize=2,
  breaklines=true,
  breakatwhitespace=true,
  aboveskip=6pt,
  belowskip=6pt,
}
\begin{document}

\title{DEAL-300K: \textbf{D}iffusion-based \textbf{E}diting \textbf{A}rea \textbf{L}ocalization with a 300K-Scale Dataset and Frequency-Prompted Baseline}

\author{
    Rui Zhang,
    Hongxia Wang, 
    Hangqing Liu,
    Yang Zhou,
    Qiang Zeng
    \thanks{This work was supported by the National Natural Science Foundation of China (NSFC) under Grant 62272331 and U25A20422, the Key Laboratory of Data Protection and Intelligent Management, Ministry of Education, Sichuan University, and the Fundamental Research Funds for the Central Universities under Grant SCU2023D008.Corresponding author: Hongxia Wang (E-mail: hxwang@scu.edu.cn).}
    \thanks{The authors are with the School of Cyber Science and Engineering, Sichuan University, Chengdu 610065, China, and also with the Key Laboratory of Data Protection and Intelligent Management, Ministry of Education, Sichuan University, Chengdu 610065, China. E-mails: zhangrui182@mails.ucas.ac.cn; hxwang@scu.edu.cn;liuhanqing0520@stu.scu.edu.cn; yzhoulv@foxmail.com, zengqiang0831@163.com. }
}

\markboth{Journal of \LaTeX\ Class Files,~Vol.~14, No.~8, August~2021}%
{Shell \MakeLowercase{\textit{et al.}}: A Sample Article Using IEEEtran.cls for IEEE Journals}

\IEEEpubid{0000--0000/00\$00.00~\copyright~2021 IEEE}


\maketitle

\begin{abstract}
Diffusion-based image editing has made semantic-level image manipulation easy for general users, but it also enables realistic local forgeries that are hard to localize. Existing benchmarks mainly focus on the binary detection of generated images or the localization of manually edited regions and do not reflect the properties of diffusion-based edits, which often blend smoothly into the original content. We present Diffusion-Based Image Editing Area Localization Dataset~(DEAL-300K), a large-scale dataset for diffusion-based image manipulation localization (DIML) with more than 300,000 annotated images. We build DEAL-300K by using a multi-modal large language model to generate editing instructions, a mask-free diffusion editor to produce manipulated images, and an active-learning change-detection pipeline to obtain pixel-level annotations. On top of this dataset, we propose a localization framework that uses a frozen Visual Foundation Model (VFM) together with Multi-Frequency Prompt Tuning (MFPT) to capture both semantic and frequency-domain cues of edited regions. Trained on DEAL-300K, our method reaches a pixel-level F1 score of $82.56\%$ on our test split and $80.97\%$ on the external CoCoGlide benchmark, providing strong baselines and a practical foundation for future DIML research.The dataset can be accessed via \url{https://github.com/ymhzyj/DEAL-300K}, and the code will be open-sourced upon acceptance.

\end{abstract}

\begin{IEEEkeywords}
Image manipulation localization, Diffusion-based image editing, Multimodal large language models.
\end{IEEEkeywords}

\section{Introduction}

\begin{figure}[thbp]
	\centering
	\includegraphics[width=\columnwidth]{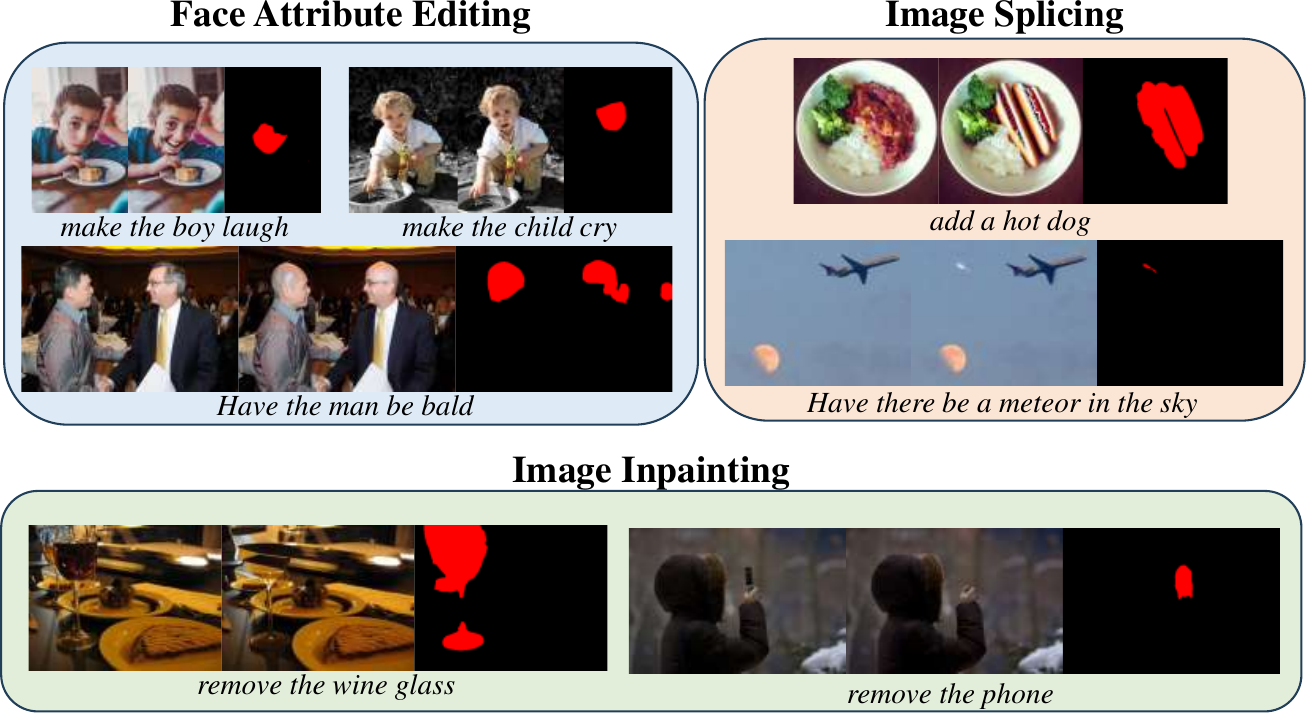}
	\caption{Sample images and annotations from the proposed DEAL-300K dataset. From left to right: the original image, its edited version, and the annotations. Below each set is the instruction used for editing. Red indicates the edited areas. }
    \label{generated_samples}
\end{figure}

The rise of Artificial Intelligence Generated Content (AIGC) technologies, especially diffusion models~\cite{DBLP:conf/nips/HoJA20, DBLP:conf/cvpr/RombachBLEO22}, has fundamentally changed digital image editing. Traditionally, tools like Adobe Photoshop required significant expertise, limiting advanced editing to professionals. Now, instruction-driven diffusion models~\cite{DBLP:conf/icml/NicholDRSMMSC22, DBLP:conf/cvpr/BrooksHE23} make complex editing accessible to a wider audience through simple language instructions, as shown in Figure~\ref{generated_samples}. While this accessibility is beneficial, it also raises concerns about the authenticity of content, particularly in the context of misinformation and digital forgeries.

Image manipulation localization (IML) has made significant progress, especially in identifying manually forged images~\cite{DBLP:journals/tcsv/LiuLCL22,DBLP:conf/cvpr/LiuSPC23,DBLP:conf/cvpr/WangWCHSLJ22}. Traditional methods often focus on detecting specific anomalies, such as digital noise~\cite{DBLP:conf/icassp/ZhouWZZM23,DBLP:conf/cvpr/0001AN19,DBLP:journals/tifs/CozzolinoV20}, which are common in manually edited images. However, these techniques struggle with diffusion-based edits, which lack distinct artifacts and maintain higher semantic and visual integrity. In the context of multimedia forensics, the development of benchmark datasets is crucial for localizing diffusion-based image editing. While research on Diffusion-based Image Manipulation Localization (DIML) is ongoing~\cite{DBLP:conf/cvpr/JiaHZJCL23, DBLP:conf/cvpr/GuillaroCSDV23, Tantaru_2024_WACV}, existing datasets remain small compared to those for traditional image manipulation detection, such as CatNet~\cite{DBLP:journals/ijcv/KwonNYLK22}, and deepfake detection, such as GenImage~\cite{DBLP:conf/nips/ZhuCYHLLT0H023}. Thus, expanding and diversifying DIML datasets is critical for improving localization algorithms and adapting to evolving manipulation techniques.

\IEEEpubidadjcol
The main challenge lies in the labor-intensive nature of annotating pixel-level diffusion-based edits. Traditional methods, such as those used in IML tasks~\cite{DBLP:conf/nips/KniazKR19, DBLP:journals/ijcv/KwonNYLK22}, often rely on simpler techniques like random object segmentation and pasting to generate synthetic datasets with automatic annotations. However, this approach lacks context and coherence, resulting in lower-quality forgeries. In contrast, diffusion-based editing generates more realistic content but requires detailed image editing instructions, making automation difficult. Moreover, because the manipulated regions are not explicitly defined, additional manual annotation is necessary to identify these regions, complicating the creation of large-scale datasets.

To address these challenges, we propose a large-scale \textbf{D}iffusion-Based Image \textbf{E}diting \textbf{A}rea \textbf{L}ocalization Dataset~(DEAL-300K), a comprehensive collection of more than 300,000 annotated images designed specifically for DIML task. Benefiting from recent advancements in multimodal large language models~(MLLMs)~\cite{DBLP:conf/icml/RadfordKHRGASAM21,DBLP:journals/corr/abs-2201-12086,DBLP:journals/corr/abs-2304-10592, cai2023vipllava}, we innovatively fine-tuned a Qwen-VL~\cite{DBLP:journals/corr/abs-2308-12966} to automatically generate high-quality editing instructions that align with the semantic information of images. Furthermore, to alleviate manual labor, we designed a data annotation process based on active learning and change detection. Our approach not only provides annotated datasets for the largest-scale DIML tasks to date but also offers a feasible method for generating even larger-scale datasets for future research.

In conjunction with the DEAL-300K dataset, we introduce a framework utilizing Visual Foundation Models (VFMs)~\cite{DBLP:conf/cvpr/FangWXSWW0WC23, DBLP:journals/corr/abs-2303-11331, DBLP:journals/corr/abs-2304-07193} as frozen encoders, combined with Multi-Frequency Prompt Tuning~(MFPT). This framework enables the capture of deep latent semantic anomalies within diffusion-based edited images, effectively leveraging the powerful prior knowledge encoded in VFMs. Additionally, by utilizing a minimal amount of frequency domain information as prompts, the framework can extract crucial low-level details that might have been overlooked, thus addressing the need for localizing regions edited by diffusion-based models.

To evaluate the robustness and efficacy of our dataset and method, we conducted comprehensive tests using various algorithms on our newly introduced benchmarks and three well-established ones~\cite{DBLP:conf/cvpr/JiaHZJCL23, DBLP:conf/cvpr/GuillaroCSDV23,Tantaru_2024_WACV}. These results highlighted the substantial challenges current localization algorithms face with our benchmarks. Our method effectively localized diffusion-edited regions, achieving state-of-the-art performance with a pixel-level F1 score of $82.56\%$ on our in-domain test set and $80.97\%$ on the external test set, CoCoGlide~\cite{DBLP:conf/cvpr/GuillaroCSDV23}. This outcome further confirms the broader value of our dataset, which can be utilized to enhance models designed to localize edits from various data sources and different diffusion models.

To summarize, our contributions are as follows:
\begin{itemize}
    \item DEAL-300K dataset is constructed with over 300,000 images for DIML task.
    \item An automated pipeline has been developed for generating editing instructions and pixel annotations, significantly reducing manual labor.
    \item A framework merging Visual Foundation Models (VFMs) with Multi-Frequency Prompt Tuning (MFPT) has been introduced, utilizing visual knowledge and frequency information to accurately localize edits made by diffusion models.
    \item Our dataset and model excel at accurately locating diffusion-based edits, establishing new benchmarks, and providing essential support for advancing localization models across various data and diffusion models.
\end{itemize}

\section{Related Work}

Recent advances in diffusion models~\cite{DBLP:conf/nips/HoJA20,DBLP:conf/cvpr/RombachBLEO22} have significantly advanced text-guided image editing algorithms. These algorithms can be categorized into two main types: mask-required methods and mask-free methods. Mask-required methods, such as Glide~\cite{DBLP:conf/icml/NicholDRSMMSC22}, DALL-E2~\cite{DBLP:journals/corr/abs-2204-06125}, and DiffEdit~\cite{DBLP:conf/iclr/CouaironVSC23}, require specifying the exact area to edit along with the text prompt. While this results in better controllability and lower failure rates, it also increases editing costs. In contrast, mask-free methods, such as NULL-text~\cite{DBLP:conf/cvpr/MokadyHAPC23}, SDEdit~\cite{DBLP:conf/iclr/MengHSSWZE22}, and InstructPix2Pix~\cite{DBLP:conf/cvpr/BrooksHE23}, rely solely on text prompts, offering greater flexibility but potentially leading to less controllable outcomes. Both approaches represent significant advances in generative image editing, with broad application potential and risks of misuse. This study focuses on datasets generated with mask-free methods, as existing DIML datasets~\cite{DBLP:conf/cvpr/GuillaroCSDV23, DBLP:conf/cvpr/JiaHZJCL23, Tantaru_2024_WACV} mainly use mask-required methods. Our results show that data generated by both types exhibit similar characteristics, suggesting that both approaches are viable for diffusion-based image editing tasks. Specifically, we utilized the InstructPix2Pix model, fine-tuned on the latest MagicBrush~\cite{DBLP:conf/nips/ZhangMCSS23} dataset. Built on the Latent Diffusion Model (LDM)~\cite{DBLP:conf/cvpr/RombachBLEO22}, it embeds image features into semantic features via a pre-trained visual encoder, interacts with the encoded text prompts, and uses Gaussian noise for denoising and decoding. This method ensures good visual and semantic consistency and is widely applied in various tasks~\cite{DBLP:journals/corr/abs-2403-13248, DBLP:journals/corr/abs-2403-14468}, making it one of the most representative algorithms in image editing.

\subsection{Image Manipulation Localization~(IML)} 
Image Manipulation Localization (IML) focuses on detecting manually edited pixels in images. Early methods concentrated on detecting anomalies such as noise~\cite{DBLP:conf/icassp/ZhouWZZM23,DBLP:conf/cvpr/0001AN19,DBLP:journals/tifs/CozzolinoV20}, frequency artifacts~\cite{DBLP:conf/iccv/ChenDJC021,DBLP:conf/cvpr/WangWCHSLJ22}, and JPEG traces~\cite{DBLP:journals/ijcv/KwonNYLK22}. Recent approaches~\cite{DBLP:conf/aaai/ZhangLC24, DBLP:conf/cvpr/GuillaroCSDV23, DBLP:conf/cvpr/LiuSPC23, DBLP:journals/tcsv/LiuLCL22} have employed advanced networks like HRNet~\cite{SunXLW19} and Segformer~\cite{DBLP:conf/nips/XieWYAAL21} to enhance performance. Li et al.~\cite{DBLP:journals/tmm/LiPZQ23} introduced a semantic refined bi-directional feature integration module to improve feature representation, while Liu et al.~\cite{10589438} proposed a boundary-guided approach. Peng et al.~\cite{DBLP:journals/tifs/PengTMLH24} explored reinforcement learning for image forgery localization. While these methods work well for traditional IML tasks, they struggle with the unique challenges posed by diffusion-based edits. Recent datasets~\cite{DBLP:conf/cvpr/GuillaroCSDV23, DBLP:conf/cvpr/JiaHZJCL23} have attempted to address this gap, but their limited scale hinders training effectiveness. The Patches dataset~\cite{Tantaru_2024_WACV} is tailored for facial image manipulation localization, yet its applicability to broader scenarios is limited.

\subsection{Deepfake detection of Diffusion-based Images} 

In forensic analysis of diffusion images, the current focus is on deepfake detection, distinguishing images generated by diffusion models from real ones. The superior visual quality of diffusion-generated images, compared to those from traditional methods, poses challenges in deepfake detection. Given the high quality of diffusion images, incorporating priors in pre-trained networks is crucial to effectively distinguish between real and synthetic content~\cite{DBLP:conf/icassp/CorviCZPNV23,DBLP:conf/iccv/WangBZWHCL23}. Yan et al.~\cite{DBLP:journals/tmm/JuJCGL24} proposed a global and local feature fusion framework for AI-Synthesized Image Detection. 
GenImage dataset~\cite{DBLP:conf/nips/ZhuCYHLLT0H023} provides a significant resource for deepfake detection, although it lacks images for DIML tasks.

\section{DEAL-300K Dataset}

This section introduces DEAL-300K, a large-scale automatically annotated dataset for Diffusion-based Image Manipulation Localization (DIML). We describe the pipeline for generating edited images and region-level labels without manual masks, present statistical analyses, and compare the dataset with existing diffusion-based editing datasets.

\subsection{MLLM-Driven Dataset Generation}

\begin{figure}[thbp]
	\centering
	\includegraphics[width=\columnwidth]{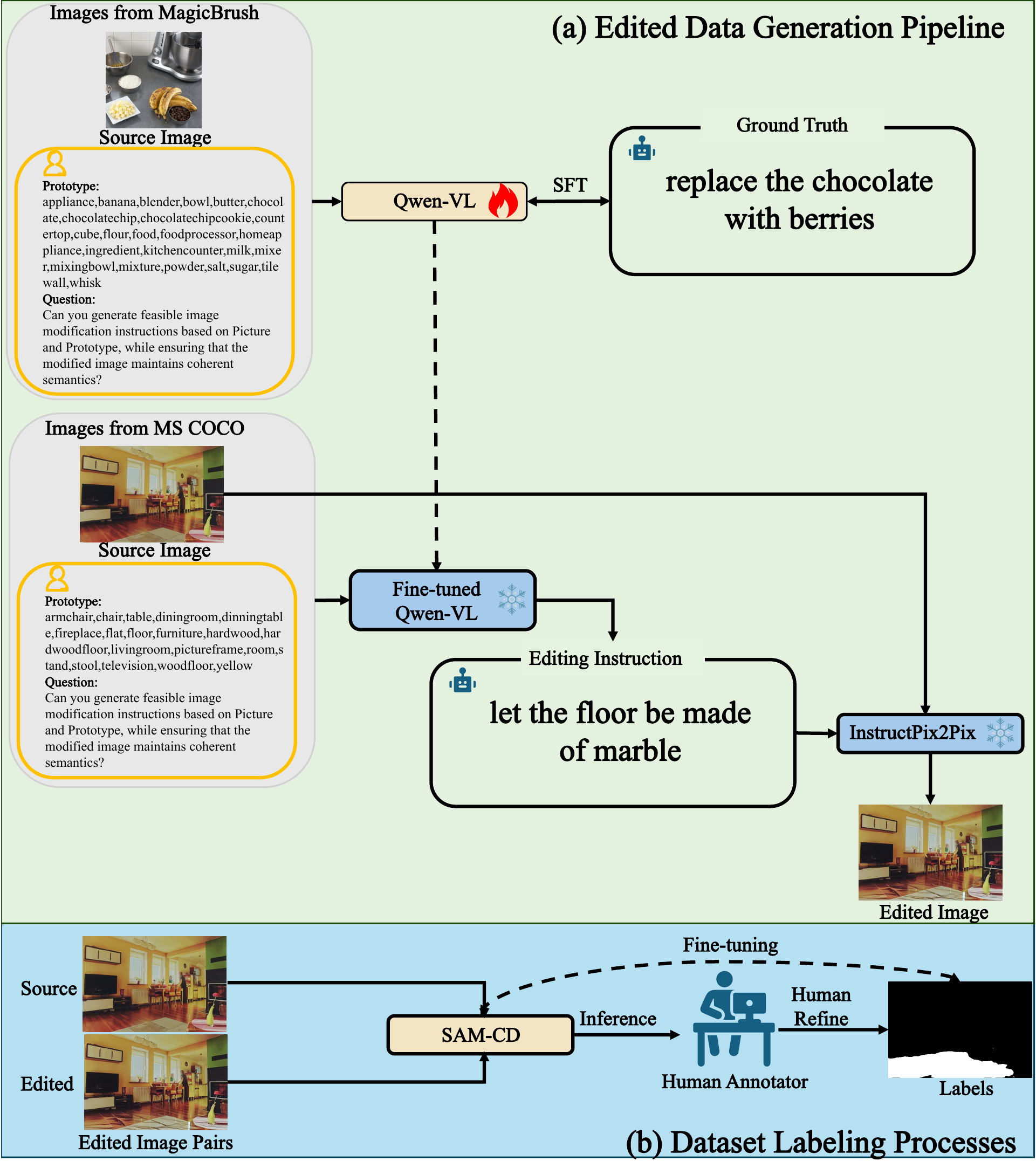}
	\caption{MLLM-driven workflow for generating edited images. SFT refers to Supervised Fine-Tuning.}
    \label{generated_pipeline}
\end{figure}

Figure~\ref{generated_pipeline}(a) illustrates our MLLM-driven pipeline. We fine-tune Qwen-VL~\cite{DBLP:journals/corr/abs-2308-12966} on MagicBrush~\cite{DBLP:conf/nips/ZhangMCSS23} so that it can generate image editing instructions from image inputs. We then apply the fine-tuned model to real images from MS COCO~\cite{DBLP:conf/eccv/LinMBHPRDZ14} to obtain image-specific instructions. Finally, InstructPix2Pix~\cite{DBLP:conf/cvpr/BrooksHE23} takes each original image and corresponding instruction to produce an edited image.

\subsubsection{\textbf{Fine-tuning Model and Generating Instructions}}

MagicBrush contains 10,388 image pairs created using DALL-E2~\cite{DBLP:journals/corr/abs-2204-06125}, including original images, edited images, instructions, and pixel-level masks. Although originally designed for text-guided image editing, its pixel-level annotations make it well suited for DIML. MagicBrush also adopts a two-stage human annotation process—creation followed by independent verification—which ensures that editing instructions are semantically accurate and practically executable. These characteristics make MagicBrush an appropriate source for training our instruction generation model.

Recent applications of LLMs in text editing~\cite{DBLP:conf/cvpr/BrooksHE23,DBLP:journals/corr/abs-2311-15308} motivate us to employ multimodal LLMs (MLLMs) to generate image-editing instructions from both images and textual prototypes. We apply supervised fine-tuning (SFT) using QLoRA~\cite{DBLP:conf/nips/DettmersPHZ23}. Specifically, we select 4,778 unedited MagicBrush images and pair each image with its corresponding instruction as ground truth.

Since CLIP-based MLLMs prioritize global semantics and often neglect fine-grained details—particularly after 4-bit quantization—we integrate RAM~\cite{DBLP:journals/corr/abs-2306-03514}, built upon SAM~\cite{DBLP:conf/iccv/KirillovMRMRGXW23}. RAM generates pixel-level object tags (e.g., color, material, pose) that serve as local visual prototypes, enhancing fine-grained cue integration. To fully leverage these tags, we enrich the training data with a structured instruction template. This template explicitly guides the model to produce semantically consistent edit instructions in a constrained format, utilizing RAM’s fine-grained annotations. The training prompt template is defined as:

\begin{lstlisting}[style=customstyle]
@user: 'Picture:<img> image path <\img> 
        Prototype: <prototypes> RAM prototypes <\prototypes>
        Generate image edit instructions in <action> <object> <attribute/position> format while maintaining semantic consistency'
@assistant: '<action> <object> <attribute/position>'
\end{lstlisting}

We fine-tune the 4-bit quantized Qwen-VL-7B using QLoRA implemented in Swift~\cite{swift}. Pretrained quantized models for large multimodal models remain scarce~\cite{DBLP:conf/nips/LiuLWL23a, zheng2023minigpt5, DBLP:conf/nips/Dai0LTZW0FH23}; directly quantizing full-precision models often yields suboptimal results when fine-tuning data are limited. The available 4-bit Qwen-VL checkpoint alleviates this issue. We find that 4-bit QLoRA performs well with structured prompts, whereas unstructured prompts cause the model to output repetitive content or plain image descriptions.

After fine-tuning, the model generates two editing instructions for each of the 123,287 images in MS COCO.

\subsubsection{\textbf{Generating Edited Images}}

Existing datasets such as AutoSplice, MagicBrush, and CoCoGlide are built using mask-required editing models (e.g., DALL-E2, GLIDE), which assume access to pixel-level masks and therefore do not reflect mask-free diffusion-based editing. Mask-free editing reduces user burden and better aligns with modern instruction-driven image editing workflows.

We adopt InstructPix2Pix, fine-tuned on MagicBrush, as our editing model due to its simplicity and ability to perform instruction-driven edits without masks. It has been adopted in downstream tasks such as MORA~\cite{DBLP:journals/corr/abs-2403-13248} and AnyV2V~\cite{DBLP:journals/corr/abs-2403-14468} for frame-level video editing, indicating its practical relevance.

Using the generated instructions and their corresponding images, InstructPix2Pix produces 246,574 edited images. Each image is edited with a randomly selected seed to improve diversity and better approximate real-world editing variability.

\subsection{Dataset Labeling with Active Learning}

\begin{figure}[thbp]
\centering
\includegraphics[width=0.8\columnwidth]{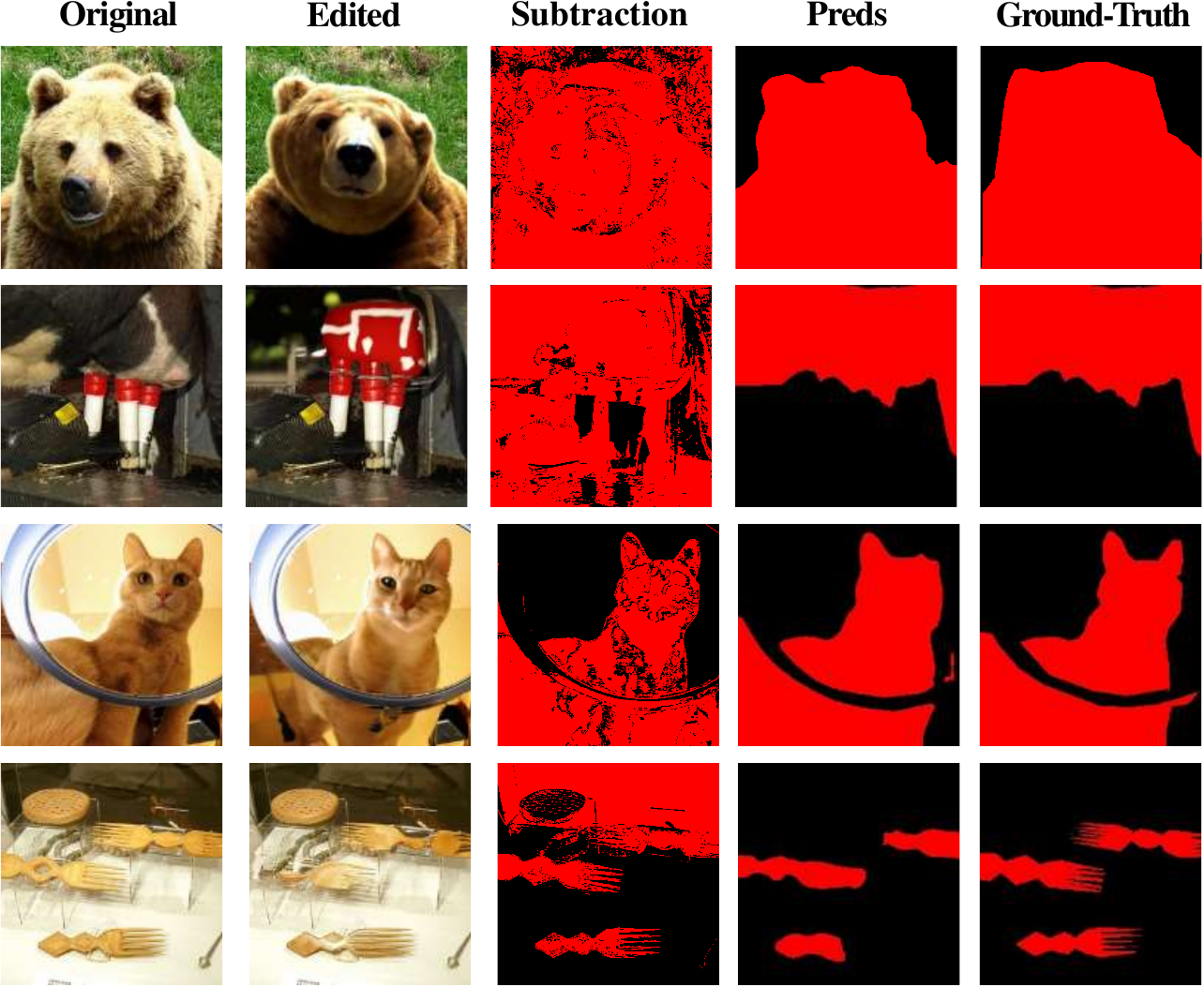}
\caption{Comparison of automated annotations with ground truth from CoCoGlide~\cite{DBLP:conf/cvpr/GuillaroCSDV23}, including subtraction analysis.}
\label{samsamples}
\end{figure}

Pixel-level annotation for mask-free edited images is challenging because diffusion models modify all image pixels slightly to maintain global photometric consistency. As shown in Figure~\ref{samsamples}, naive image subtraction yields extensive false positives. We therefore reformulate annotation as semantic change detection. Our approach achieves 94.69\% pixel accuracy on CoCoGlide at threshold 0.5, approaching human-level quality.

Inspired by active learning–based labeling~\cite{DBLP:conf/accv/XieFCSMS20,DBLP:conf/cvpr/SiddiquiVN20,DBLP:conf/eccv/GaoZYADP20}, we develop an interactive annotation pipeline (Figure~\ref{generated_pipeline}(b)). Since general-purpose change detection models do not exist, we re-train SAM-CD~\cite{DBLP:journals/tgrs/DingZPTYB24} from scratch on MagicBrush image pairs.

In the first stage, we annotate 1,000 edited–original image pairs using SAM-CD with a conservative threshold of 0.5. Each mask is corrected using ISAT~\cite{ISAT_with_segment_anything}. We perform four rounds of fine-tuning on iteratively corrected masks (1,000 pairs each), yielding 4,000 refined labeled pairs. We manually discard 401 clearly low-quality edits, resulting in 3,599 edited images and 1,901 original images (5,500 total), which constitute the DEAL-300K test set. This stage requires approximately 30 hours of work by a single annotator.

In the second stage, we process 2,000 additional real images and their edited versions. With a threshold of 0.5, the refined SAM-CD achieves 90.1\% pixel accuracy on CoCoGlide. We adopt the following rules:  
(1) samples with mean prediction probability $>$ 0.5 are accepted;  
(2) samples in $[0.3,0.5]$ are manually checked;  
(3) samples $<$ 0.3 are discarded.  
Images with predicted areas $>$99\% or $<$1\% are also removed. This yields 2,333 edited images and 1,656 real images as the validation set. We also ensure that each real image and all of its edited variants appear in only one of train/val/test splits to prevent data leakage. This stage requires about 7 hours of manual effort.

We evaluate the final model on CoCoGlide under multiple thresholds (Table~\ref{tab:samcd_cocoglide}). At a threshold of 0.1, performance remains high and manual effort is minimized. We therefore adopt 0.1 for labeling the remaining data. Random checks (10\%) confirm that annotation noise remains low.

\begin{table}[htbp]
\centering
\caption{Labeling performance of SAM-CD at different thresholds on CoCoGlide.}
\label{tab:samcd_cocoglide}
\resizebox{0.6\columnwidth}{!}{%
\begin{tabular}{cccc}
\hline
Threshold & pF1(\%)        & IoU(\%)        & pACC(\%)       \\ \hline
0.1       & \textbf{91.74} & \textbf{89.37} & \textbf{95.73} \\
0.3       & 90.42          & 88.31          & 95.39          \\
0.5       & 88.61          & 86.42          & 94.69          \\ \hline
\end{tabular}%
}
\end{table}

Using this setting and discarding samples with $<$1\% or $>$99\% predicted mask coverage, we obtain the DEAL-300K training set: 115,814 original images and 21,165 edited images. Although these are not individually verified, spot checks confirm overall quality, and prior work~\cite{DBLP:conf/cvpr/XieWWWT16,DBLP:conf/iclr/WeiSCM21,DBLP:conf/nips/MullerKH19} shows that modest label noise can improve robustness in real-world scenarios.

In total, annotating the 300K-scale dataset requires approximately 42 hours of manual effort by one person. The process scales well and can be parallelized in industrial settings. We release the trained SAM-CD model to support similar annotation pipelines.

\subsection{Visualization of the Dataset Generation Pipeline}

\begin{figure*}[htbp]
\centering
\includegraphics[width=1.5\columnwidth]{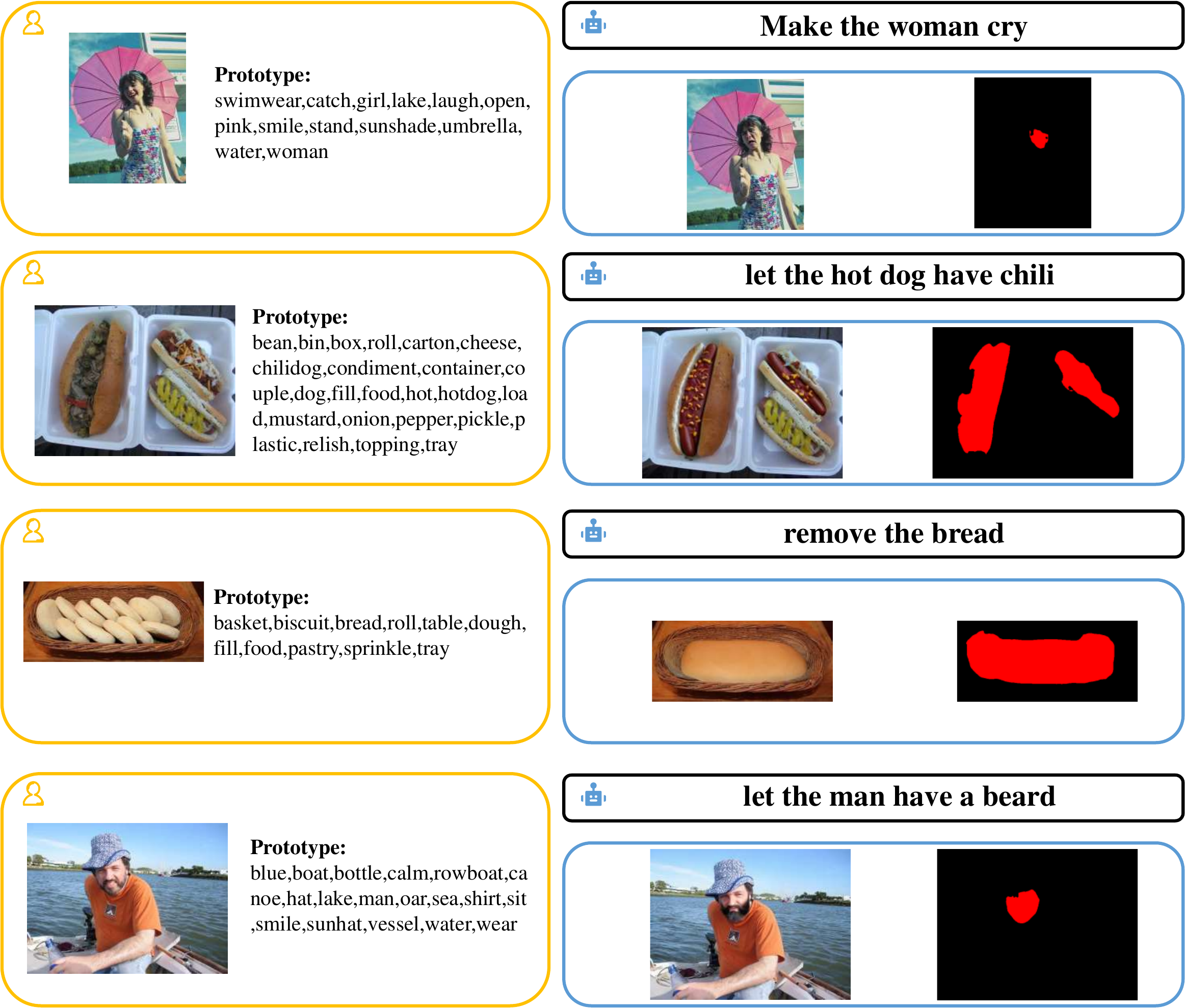}
\caption{Examples from the DEAL-300K generation pipeline. Qwen-VL generates editing instructions, InstructPix2Pix produces edited images, and SAM-CD generates pixel-level masks.}
\label{datasetpipelinevisual}
\end{figure*}

Figure~\ref{datasetpipelinevisual} visualizes prototypes, generated instructions, edited images, and annotations in our pipeline. Since the training set labels are produced automatically, these examples illustrate that the pipeline can generate reasonable annotations without human intervention.

\begin{figure}[htbp]
\centering
\includegraphics[width=\columnwidth]{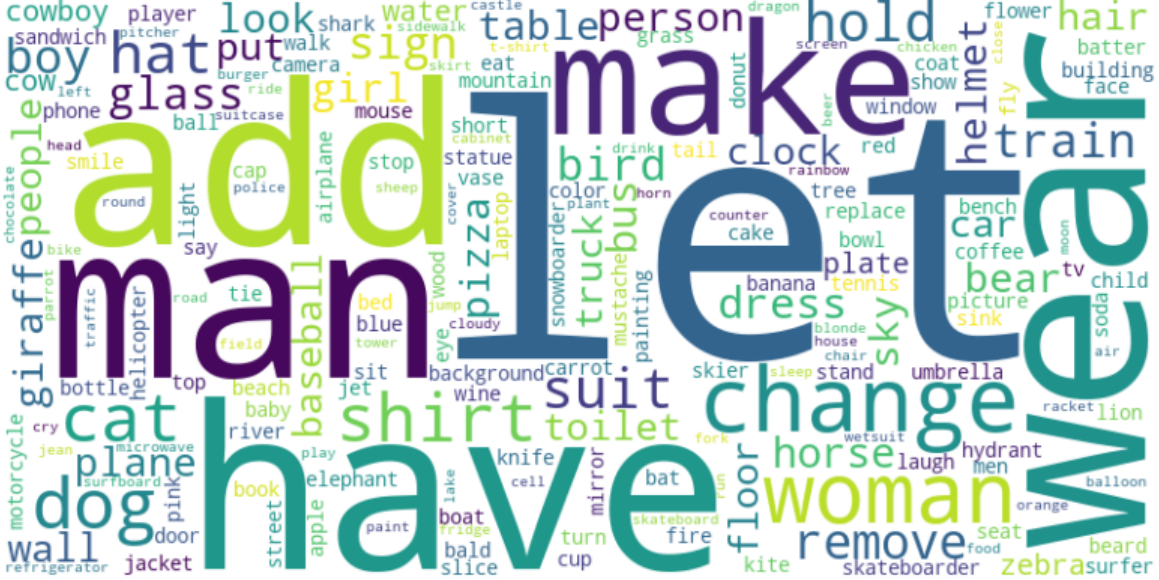}
\caption{Word cloud of editing instructions.}
\label{Wordcloud}
\end{figure}

Figure~\ref{Wordcloud} shows a word cloud of the editing instructions, indicating that our dataset covers a broad range of manipulations involving humans, animals, and objects. These include classical operations such as ``add'' and ``remove'' as well as attribute changes common in Deepfake-style editing. The diversity of operations is useful for evaluating DIML methods under varied editing scenarios.

\subsection{Analysis of Failed Samples in Editing Processes}

\begin{figure}[thbp]
\centering
\includegraphics[width=\columnwidth]{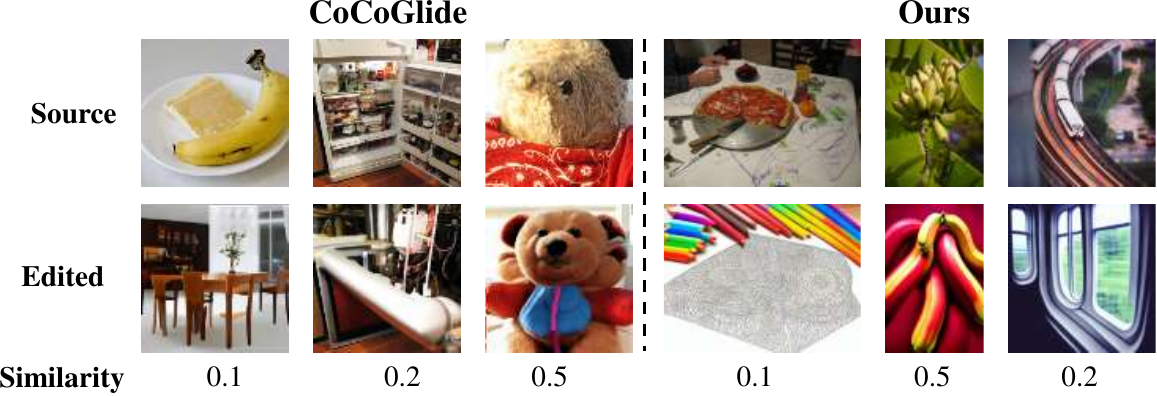}
\caption{Examples of editing failures from CoCoGlide and our generated data.}
\label{Post_processing_image}
\end{figure}

\begin{figure}[thbp]
\centering
\includegraphics[width=\columnwidth]{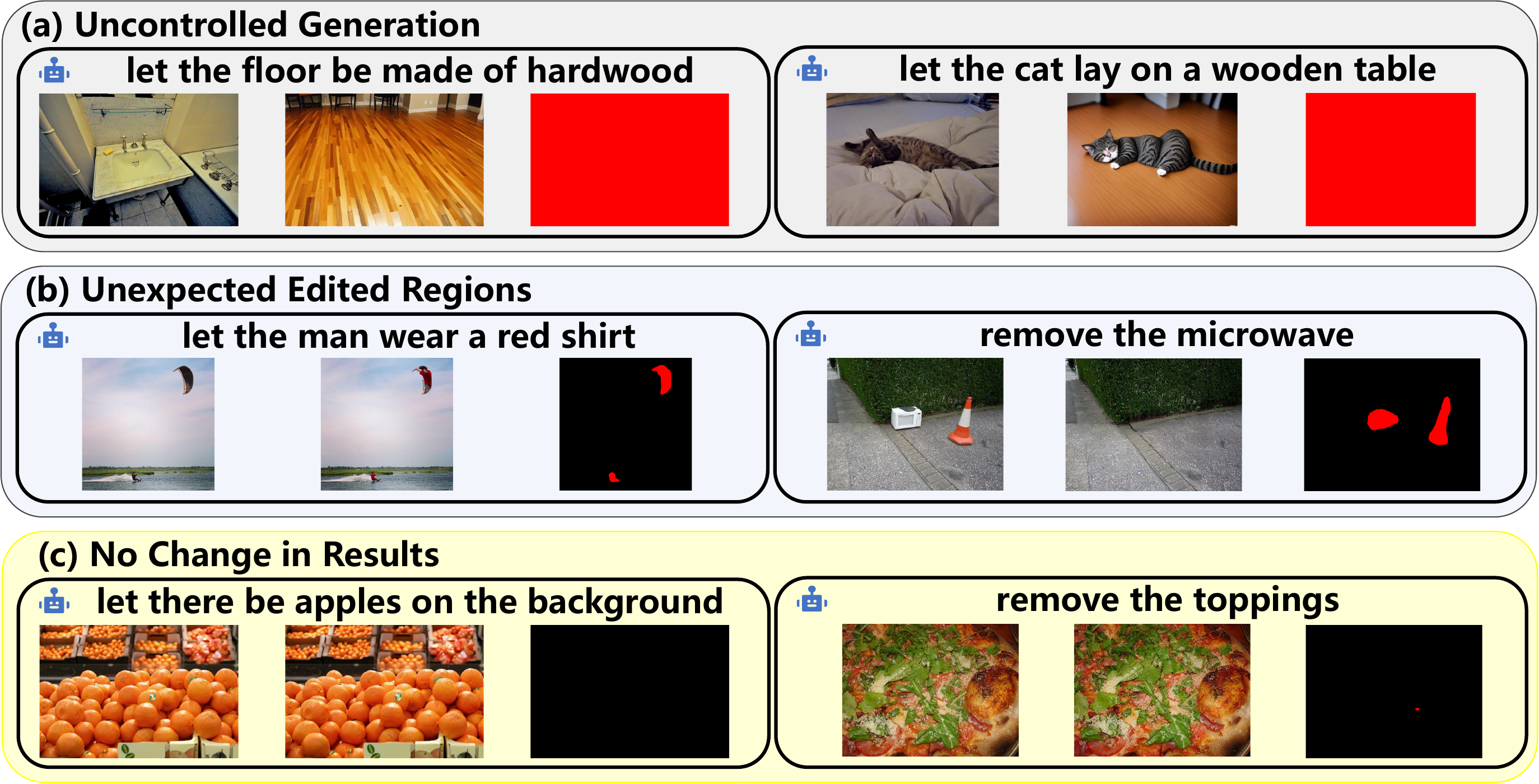}
\caption{Representative examples from the three failure categories, along with SAM-CD predictions.}
\label{failed_samples}
\end{figure}

Diffusion-based editing may suffer from cross-attention leakage or incomplete attention activation~\cite{DBLP:conf/cvpr/MokadyHAPC23, DBLP:conf/nips/0060YYB023, DBLP:conf/iclr/CouaironVSC23, DBLP:conf/cvpr/KawarZLTCDMI23}, producing unrealistic or unintended edits. Because the goal of DIML is to localize manipulations rather than maximize visual fidelity, we categorize failure modes based on predicted forged area:

\begin{enumerate}
    \item \textbf{Uncontrolled generation ($\geq$99\% area).} The model generates almost entirely new content. These samples represent global synthesis and are removed from train/val/test. SAM-CD can identify such samples (Figure~\ref{failed_samples}(a)).
    \item \textbf{Unexpected edits ($<99\%$ area).} The model edits unintended regions. These samples still contain meaningful manipulation traces. We retain them for training but remove low-quality cases from validation and test. SAM-CD accurately labels these edits (Figure~\ref{failed_samples}(b)).
    \item \textbf{No-change cases ($<1\%$ area).} The model reconstructs the image without editing. These are removed entirely. SAM-CD identifies them reliably (Figure~\ref{failed_samples}(c)).
\end{enumerate}

We evaluate the visual quality using FID~\cite{DBLP:conf/nips/HeuselRUNH17} and MUSIQ~\cite{DBLP:conf/iccv/KeWWMY21}. Compared with the manually verified MagicBrush dataset, DEAL-300K achieves lower FID and higher MUSIQ scores (Table~\ref{quality}).

\begin{table}[htbp]
\centering
\caption{Quality evaluation of edited images.}
\resizebox{0.5\columnwidth}{!}{%
\begin{tabular}{ccc}
\hline
Dataset       & FID$\downarrow$ & MUSIQ$\uparrow$ \\ \hline
MagicBrush          & 2.89        & 44.75  \\
DEAL-300K           & \textbf{2.23} & \textbf{69.17} \\ \hline
\end{tabular}%
}
\label{quality}
\end{table}

\subsection{Dataset Summary}

\begin{table*}[htbp]
\centering
\caption{Comparison of diffusion-based editing datasets.}
\label{tab:compare_datasets}
\resizebox{1.8\columnwidth}{!}{%
\begin{tabular}{ccccccc}
\hline
Dataset              & Year & Source Images & Edited Images & Image Size                    & Scenario & Generative Model       \\ \hline
CoCoGlide~\cite{DBLP:conf/cvpr/GuillaroCSDV23}            & 2023 & 512           & 512           & $256\times256$   & General  & GLIDE (Mask-Required)   \\
AutoSplice~\cite{DBLP:conf/cvpr/JiaHZJCL23}      & 2023 & 2,273  & 3,621  & $256\times256$--$4232\times4232$ & General & DALL-E2 (Mask-Required)     \\
MagicBrush~\cite{DBLP:conf/nips/ZhangMCSS23}           & 2023 & 5,313         & 10,388 (9,335 annotated)         & $1024\times1024$ & General  & DALL-E2 (Mask-Required) \\
Repaint-P2/CelebA-HQ~\cite{Tantaru_2024_WACV} & 2024 & 10,800        & 41,472        & $256\times256$   & Face     & Repaint (Mask-Required) \\ \hline
DEAL-300K (Ours) & -- & 119,371 & 221,097 & $128\times512$--$512\times576$   & General & InstructPix2Pix (Mask-Free) \\ \hline
\end{tabular}%
}
\end{table*}

DEAL-300K contains over 300K images: 330,979 for training, 3,989 for validation, and 5,500 for testing. The test set is further divided into three subsets: DEAL-A (authentic-only), DEAL-E (edited-only), and DEAL-Full (combined). DEAL-A evaluates false positives on real images, DEAL-E measures localization accuracy on manipulated images, and DEAL-Full assesses performance under mixed scenarios.

Since our generated images contain mostly single-turn edits, most have only one edited region. To evaluate cross-method and multi-turn generalization, we construct an auxiliary test set, DEAL-MB, based on MagicBrush multi-step edits. Multi-step masks are merged to create images with multiple manipulated regions (9,335 images total).

\begin{figure}[thbp]
    \centering
    \captionsetup[subfloat]{font=scriptsize}
    \subfloat[DEAL-Train]{\includegraphics[width=0.45\columnwidth]{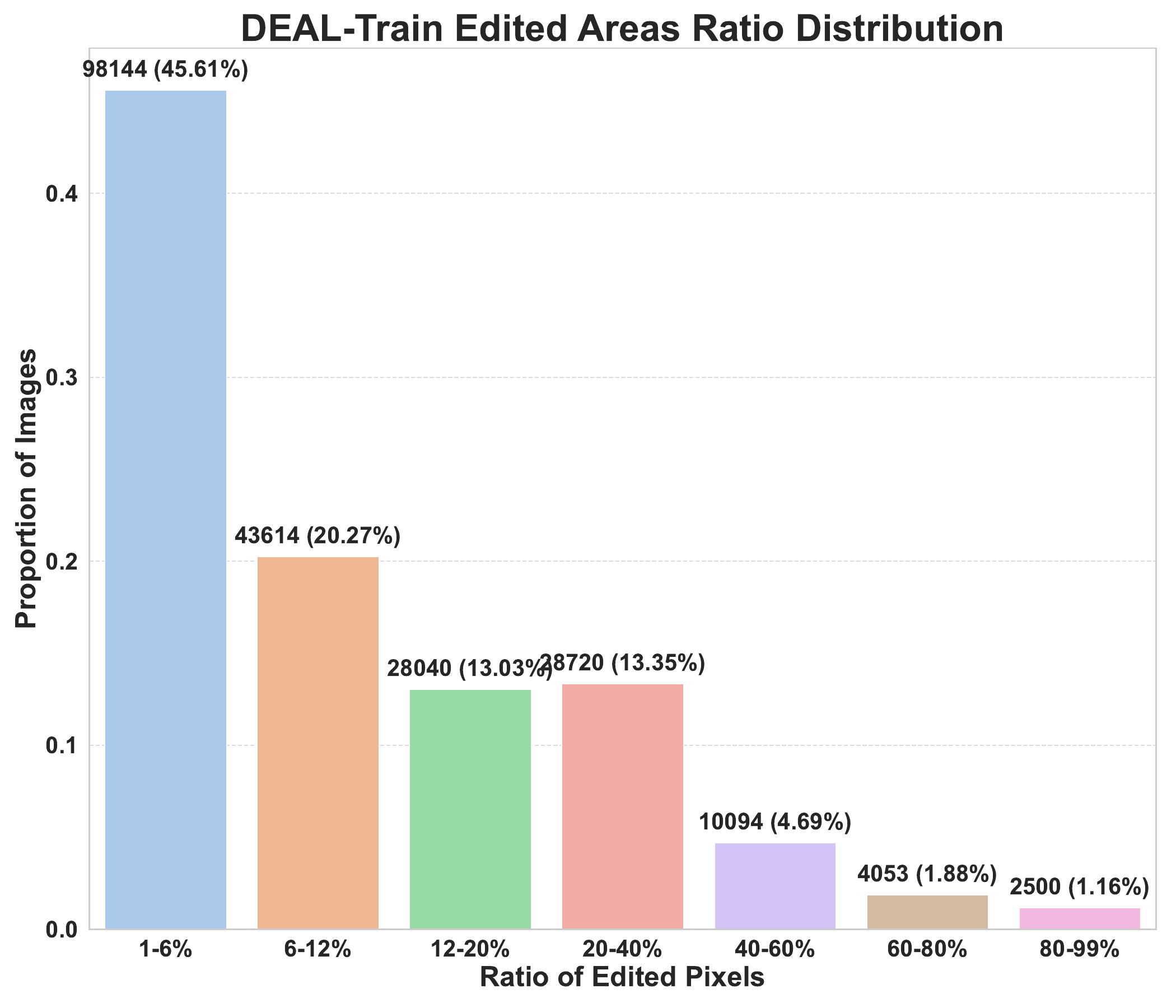}}%
    \subfloat[DEAL-Val]{\includegraphics[width=0.45\columnwidth]{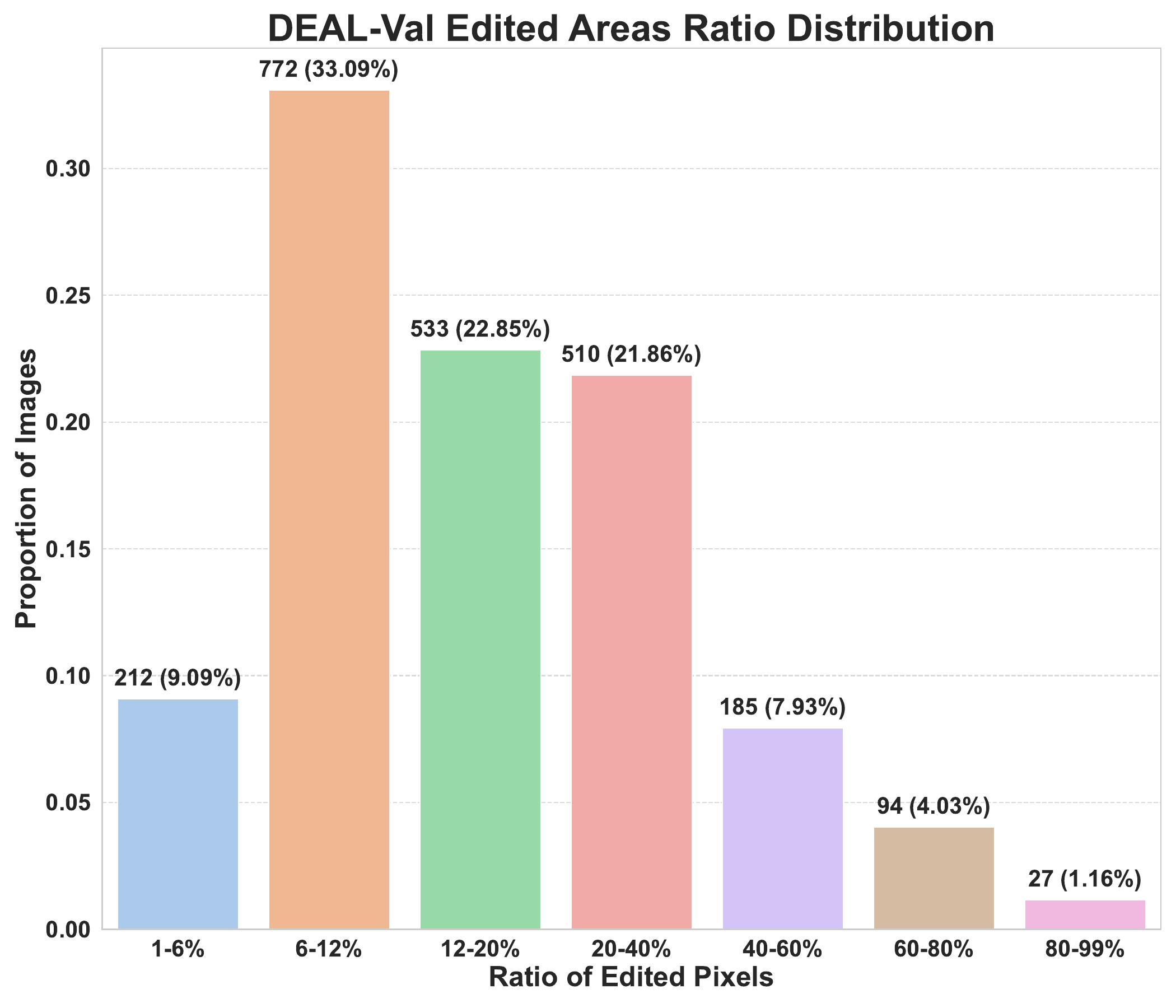}}%
    \vfil
    \subfloat[DEAL-E]{\includegraphics[width=0.45\columnwidth]{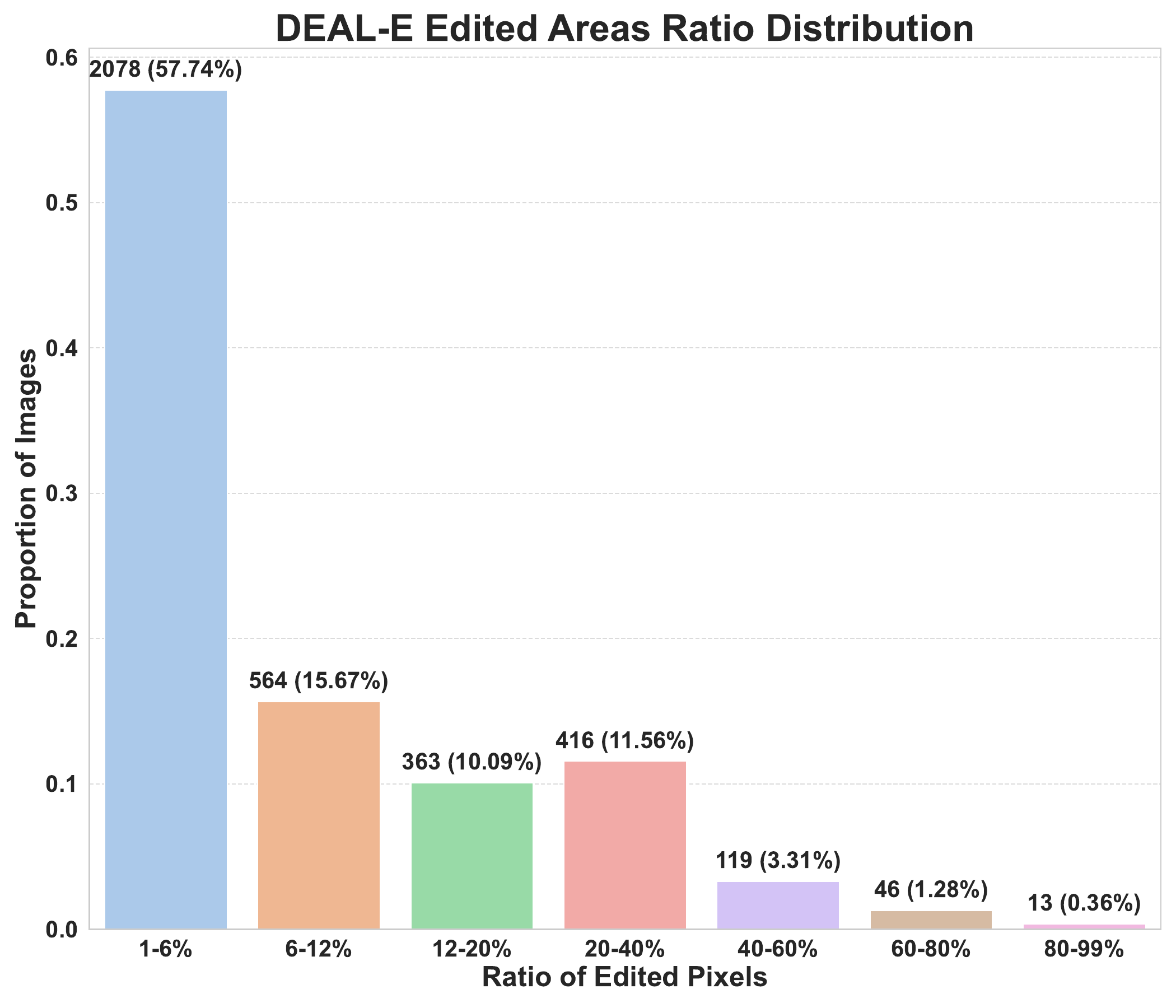}}%
    \subfloat[DEAL-MB]{\includegraphics[width=0.45\columnwidth]{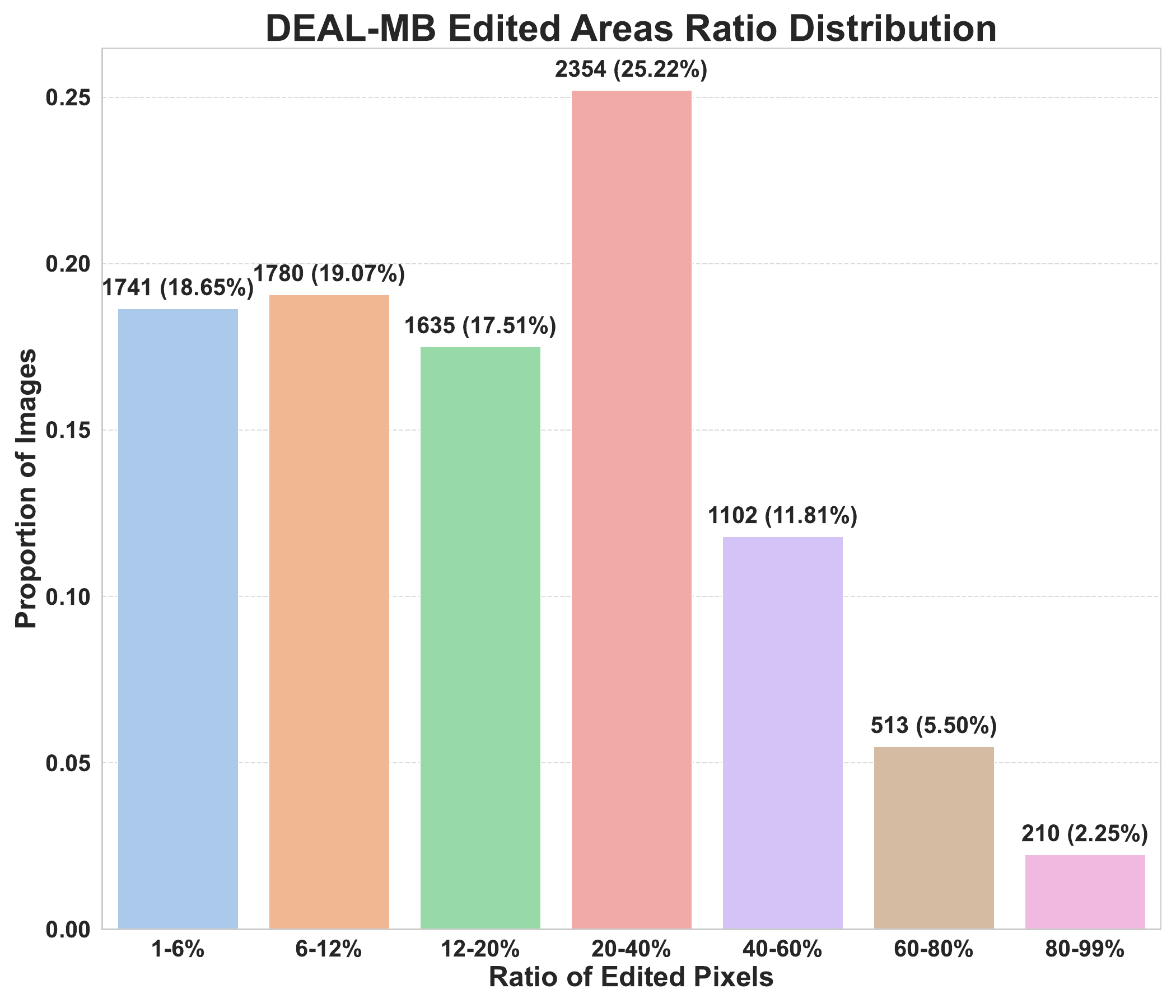}}%
    
    \caption{Distribution of edited-area ratios across DEAL-Train, DEAL-Val, DEAL-E, and DEAL-MB.}
    \label{fig:deal-distribution}
\end{figure}

Figure~\ref{fig:deal-distribution} shows that DEAL-E is dominated by small edited regions (mostly $<$12\%). Detecting such regions is challenging and representative of practical online manipulations. DEAL-Train follows a similar skewed distribution so that training emphasizes small, subtle manipulations, while DEAL-Val is more balanced to prevent overfitting to a specific edit size. DEAL-MB contains larger edits due to multi-step editing, making it suitable for evaluating generalization across manipulation scales.

\section{Proposed Approach}

\subsection{Preliminary}
Our goal is to accurately locate areas within images edited by diffusion models, a task known as DIML. Diffusion-edited images often maintain local visual and content consistency, increasing the difficulty of distinguishing edited regions, especially when using simple end-to-end model training. While larger VFMs~\cite{DBLP:conf/cvpr/FangWXSWW0WC23, DBLP:journals/corr/abs-2303-11331, DBLP:journals/corr/abs-2304-07193}, benefiting from prior knowledge of visual information, have demonstrated significant advancements in computer vision tasks, their full-parameter fine-tuning requires extensive computational resources and risks catastrophic forgetting\cite{pmlr-v234-zhai24a}. Inspired by the success of parameter-efficient fine-tuning (PEFT) in natural language processing~\cite{DBLP:conf/iclr/HuSWALWWC22, DBLP:journals/corr/abs-2309-14717, DBLP:conf/icml/HoulsbyGJMLGAG19} and computer vision~\cite{DBLP:conf/eccv/JiaTCCBHL22, DBLP:journals/corr/abs-2312-04265, DBLP:conf/aaai/MouWXW0QS24}, we introduce the Multi-Frequency Prompt Tuning (MFPT) architecture. Motivated by observable differences in the frequency domain of diffusion-generated images~\cite{DBLP:conf/icassp/CorviCZPNV23} and the lack of such specialized knowledge in existing VFMs, our innovative architecture implements prompt learning not only at the image and feature levels but also in the frequency domain. This approach enables VFMs to better specialize in localizing forged regions within diffusion-based edited images.

\subsection{Multi-Frequency Prompt Tuning Architecture}
\begin{figure*}[thbp]
	\centering
	\includegraphics[width=1.5\columnwidth]{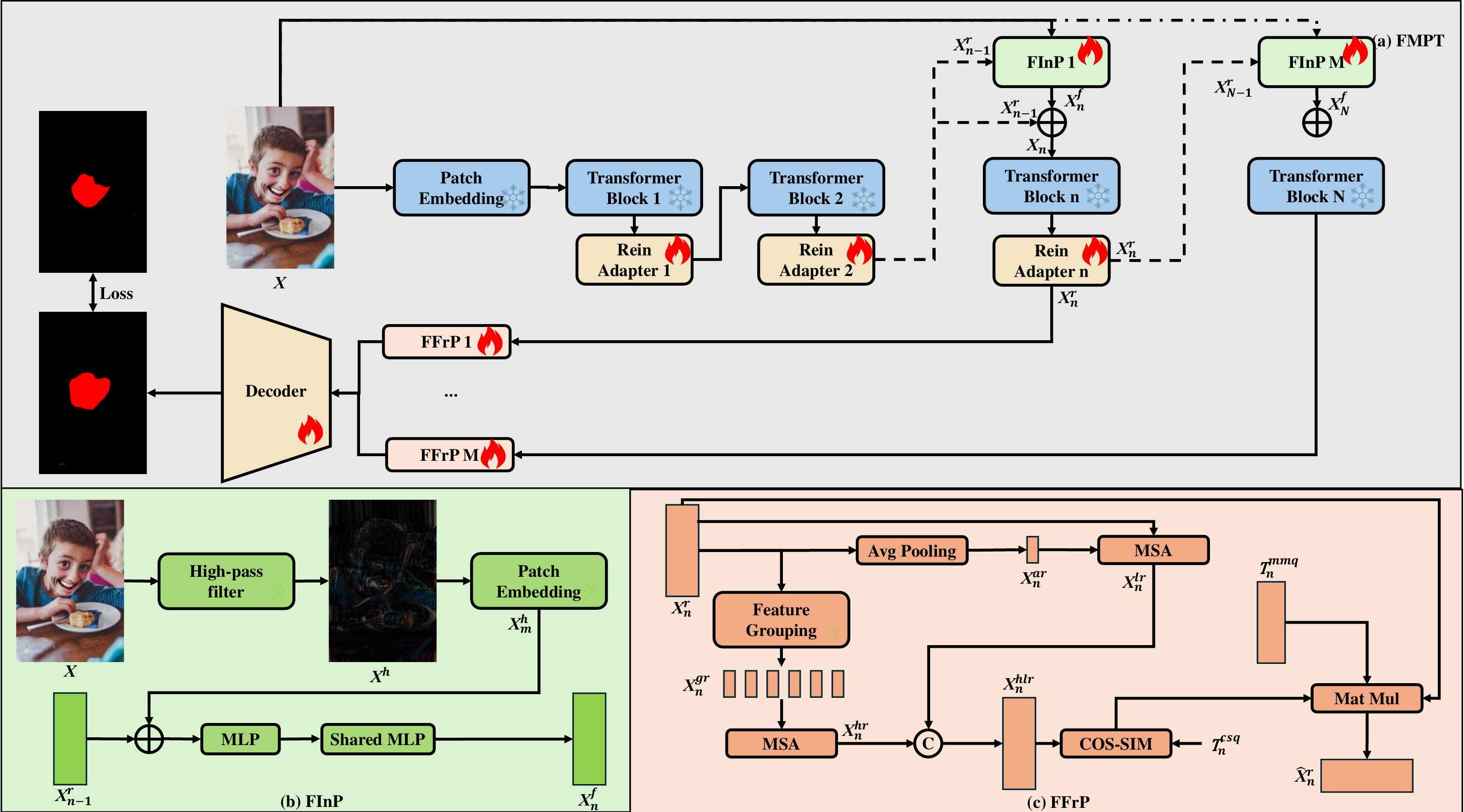}
	\caption{An overview of the introduced Multi-Frequency Prompt Tuning~(MFPT) Architecture.}
        \label{framework}
\end{figure*}

The overview of our proposed MFPT is illustrated in Figure~\ref{framework}. Building upon the foundation of Mask2Former~\cite{DBLP:conf/cvpr/ChengMSKG22} and Rein~\cite{DBLP:journals/corr/abs-2312-04265}, our model architecture is further enhanced by integrating two critical components: Frequency Input Prompters~(FInP) and Feature Frequency Prompters~(FFrP). In essence, our approach can be implemented with any variant of VFM serving as an encoder. We keep the weights of the original VFM fixed and focus on training a small subset of parameters devised by us. This strategy encompasses a diverse array of prompting mechanisms at both the input and output feature stages within the VFMs, significantly bolstering overall performance.
Each component is elaborated on in the following section.

\subsubsection{\textbf{Frequency Input Prompters}}
Drawing from prior works~\cite{DBLP:conf/cvpr/LiuSPC23, DBLP:journals/corr/abs-2401-02326}, we introduce a method termed FInP, which integrates high-frequency image features with semantic features extracted by the encoder at various output stages. This approach is inspired by the behavior of existing VFMs, which, after extensive pre-training, often prioritize global image features while overlooking lower-level texture details to maintain model robustness. However, in our specific task of localizing editing regions, influenced by the characteristics of diffusion models, the edited regions frequently retain strong visual characteristics in terms of image content but may lack fine local texture information. Therefore, harnessing high-frequency features, which capture low-level texture information, emerges as a natural choice.

Our approach utilizes a pre-trained VFM encoder, comprising $N$ transformer blocks. To improve localization precision, feature maps from $M$ distinct stages are processed by a decoder. We introduce $M$ FInP modules to enhance transformer blocks with detailed prompts based on feature input. Specifically, at the $m$-th output feature stage, corresponding to the $n$-th stage of the encoder, an input image $X \in \mathbb{R}^{W\times H \times 3}$, where $W$ is the width of the image and $H$ is the height of the image, is initially converted to grayscale by the FInP module. Subsequently, a two-dimensional Fast Fourier Transform (FFT) is applied, generating a Fourier spectrum $F$. A high-pass filter further isolates the high-frequency spectrum $F_{h}$. Following an inverse FFT on $F_{h}$, we obtain a prompt image $X^h \in \mathbb{R}^{W\times H \times 1}$, distinguished by its detail preservation and edge enhancement. This prompt image is transformed into feature vectors $X^{h}{m}$ using patch embedding from~\cite{DBLP:journals/corr/abs-2304-07193},where $C$ denotes the channel count and $L$ the sample size, ensuring compatibility with the dimensions of features $X^{r}_{n-1}$ from the $(n-1)$-th Rein Adapter from the previous stage. Subsequently, we derive the prompt feature $X^f_{n}\in \mathbb{R}^{C \times L}$ as:
\begin{equation}
X^f_{n} = f_{s}(f_{m}(X^{h}_{m} +X^{r}_{n-1})).
\end{equation}
In this equation, $f_{m}$ represents the multilayer perceptron~(MLP) at the $m$-th FInP, which integrates the features of the previous stage with frequency domain prompts and maps them to a unified feature space. Using a shared MLP, $f_{s}$, across all FInP modules, these are then transcribed into frequency domain prompts. By integrating these enriched features $X^f_{n}$ with the foundational features $X^{r}_{n-1}$ as $X_{n}$, the ensemble progresses to additional processing stages.

\subsubsection{\textbf{Feature Frequency Prompters}}
In this section, we describe enhancing features at the $m$-th output stage, related to the $n$-th transformer block.
The features $X_n$ are fed into the  $n$-th transformer block with a Rein Adapter used to adjust inputs with frequency domain prompts, producing enhanced features $X^r_n \in \mathbb{R}^{\hat{C} \times \hat{L}}$. Here, $\hat{C}$ and $\hat{L}$ indicate new feature dimensions and sample size. Our next goal is to amplify the high-frequency details in $X^r_n$ to improve the decoder’s focus on local content details. Inspired by dual attention mechanisms~\cite{DBLP:conf/nips/Pan0Z22, DBLP:conf/aaai/XuXWXGZZ24}, our FFrP employs two branches to selectively enhance both high- and low-frequency information in $X^r_n$, with a particular focus on enriching high-frequency content for improved local detail representation.

Firstly, to emphasize the importance of local features, we partitioned the feature $X^r_n$ of length $\hat{L}$ into non-overlapping groups, each consisting of samples with a length of $\hat{L}_g$. The resulting grouped samples $X^{gr}_n$ are then fed into a multi-head self-attention~(MSA) module to obtain multiple sets of high-frequency features. These features are then reconstructed in the order of the groups to restore high-frequency attention features $X^{hr}_n$ with the same scale as $X^r_n$. 

Our next step focuses on extracting low-frequency information, indicative of global semantic content. This is achieved by downsampling $X^r_n$ using average pooling, with a window scale matching the previous segmentation length $\hat{L}_g$. This process enables us to obtain features that primarily emphasize the global aspects of features.  These downsampled features then serve as keys and values in another MSA module, with $X^r_n$ fulfilling the query role. This MSA processing yields low-frequency attention features $X^{lr}_n$, scaled to match $X^r_n$. It is worth mentioning that, to reflect the difference in the importance of high-frequency and low-frequency information, the number of MSA heads in two branches differs. With a total of $N_h$ heads, the number of heads in the high-frequency branch is $r \times N_h$, while the number of heads in the low-frequency branch is $(1-r) \times N_h$, where $r \in (0.5,1]$. The number of feature channels in the two branches is also allocated in the same proportion $r$.Following this, we concatenate the high and low-frequency features along the channel dimension, creating a new feature set $X^{hlr}_n \in \mathbb{R}^{\hat{C} \times \hat{L}}$. In addition, inspired by the widespread use of the [CLS] token in VFMs,  we incorporate two learnable tokens specifically for frequency domain analysis, acting effectively as a frequency domain filter for features.  These are the frequency filter token $T^{frf}_m \in \mathbb{R}^{\hat{C}}$ and the frequency matching token $T^{frm}_m \in \mathbb{R}^{\hat{C}\times\hat{C}}$, designed to filter and align significant features from a frequency perspective. The final features, ready for the decoder, are computed as follows:
\begin{equation}
\hat{X}^{r}_n = T^{frm}_m \cdot (\text{cos\_sim}(X^{hlr}_n, T^{frf}_m)) \cdot X^r_n,
\end{equation}
where $\text{cos\_sim}(\cdot, \cdot)$ denotes the cosine similarity, and the resulting $\hat{X}^{r}_n \in \mathbb{R}^{\hat{C} \times \hat{L}}$.

\subsection{Training Losses}
During training, our model generates predictions \( P \in \mathbb{R}^{2 \times W \times H} \) and compares them to ground truth templates \( M \in \mathbb{R}^{1 \times W \times H} \). The loss function combines Dice loss and binary cross-entropy (BCE) loss to optimize the model’s performance.

Dice loss measures the overlap between predictions and ground truth, making it effective for segmentation tasks, especially with imbalanced data. BCE loss, widely used in binary classification, evaluates the pixel-wise difference between predicted probabilities and true labels.

Balancing coefficients \( \lambda_{\text{Dice}} \) and \( \lambda_{\text{CE}} \) control the contribution of each loss term:
\[
\text{Loss} = \lambda_{\text{Dice}} \cdot \text{DiceLoss}(P, M) + \lambda_{\text{CE}} \cdot \text{BCELoss}(P, M).
\]
By default, both coefficients are set to 1, allowing equal contribution from each loss. These parameters can be adjusted based on task-specific needs to optimize the model’s performance.

\section{Experimental}
\subsection{Experiment Setup}
\subsubsection{\textbf{Datasets}} We conduct experiments on three widely used benchmarks: CoCoGlide~\cite{DBLP:conf/cvpr/GuillaroCSDV23}, AutoSplice~\cite{DBLP:conf/cvpr/JiaHZJCL23}, and Repaint-P2/CelebA-HQ~\cite{Tantaru_2024_WACV}, which cover both general and facial contexts. Additionally, we introduce a new dataset, DEAL-300K dataset, which contains a diverse set of manipulated images across various contexts. DEAL-300K includes not only images with simple manipulations, such as cropping or splicing, but also complex scenarios involving multiple rounds of editing. This diversity enables a comprehensive assessment of model performance under different image manipulation conditions.

\subsubsection{\textbf{Evaluation Metrics}} Following most previous work about IML and DIML, we evaluated the localization results using pixel-level F1~(pF1), pixel-level accuracy~(pACC) and intersection over union~(IoU).
\subsubsection{\textbf{Implementation Details}} For the majority of our experiments, DINOv2~\cite{DBLP:journals/corr/abs-2304-07193} is utilized as the representative for VFM as the encoder. Our model was implemented using PyTorch~\cite{DBLP:conf/nips/PaszkeGMLBCKLGA19}. All our code runs on two RTX3090 GPUs. For data labeling, we fine-tuned the Qwen-VL and re-train SAM-CD models. Qwen-VL was fine-tuned with a batch size of 8 for 5 epochs at a learning rate of 0.0001. The batch size for SAM-CD was 16. The learning rate during the cold start phase was 0.1, following \cite{DBLP:journals/tgrs/DingZPTYB24}. The fine-tuning learning rate for the subsequent phases was 0.001, and each phase was trained for 50 epochs. We selected the model that performed the best in the validation set for subsequent tasks. For DIML tasks, our architecture used a learning rate of 0.0001 with the AdamW optimizer.  Furthermore, we replicated code for traditional IML models~\cite{DBLP:journals/tcsv/LiuLCL22,DBLP:conf/cvpr/LiuSPC23}, and a state-of-the-art DIML algorithm~\cite{Tantaru_2024_WACV}. Moreover, to the best of our knowledge, most of the base models used in existing IML tasks are typically derived from classic semantic segmentation models such as HRNet~\cite{SunXLW19} and Segformer~\cite{DBLP:conf/nips/XieWYAAL21}. These two models are particularly noteworthy as they represent two significant eras in semantic segmentation: the traditional convolutional network era and the modern Transformer-based era, respectively. Therefore, we have also re-implemented these two algorithms to provide a baseline performance comparison. All models including us were trained with a batch size of 8 and 41372 iterations, and the model with the best validation performance was saved for each algorithm to ensure a fair comparison.

\begin{figure*}[thbp]
	\centering
	\includegraphics[width=1.5\columnwidth]{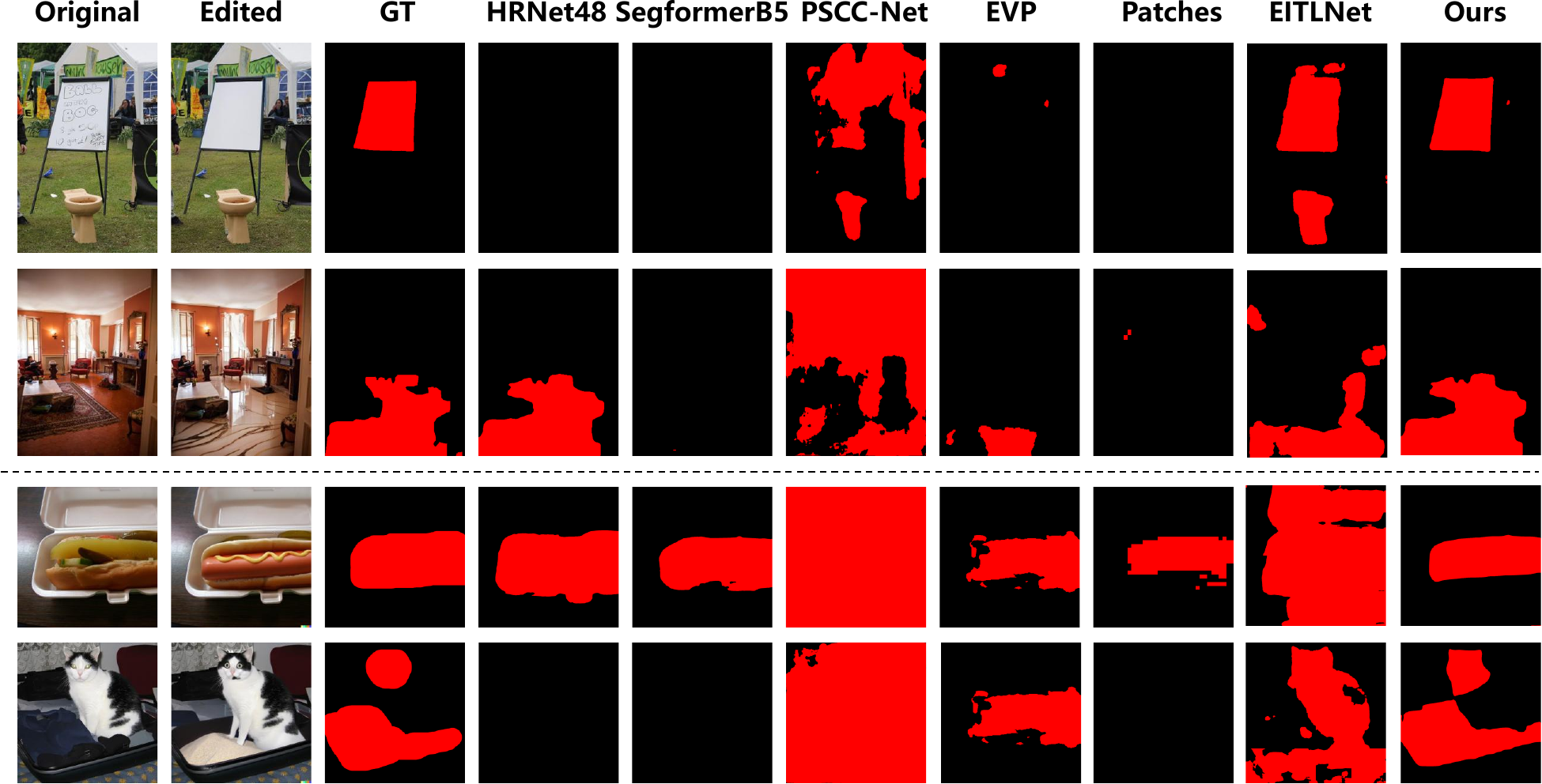}
	\caption{Qualitative Localization results on the DEAL-300K test set. The top two sets are from DEAL-E, while the bottom two sets are from DEAL-MB. Red indicates the edited areas.}
    \label{Localization_results}
\end{figure*}

\subsection{Model Training and Performance Evaluation on DEAL-300K}
To comprehensively evaluate the effectiveness of the DEAL-300K dataset and our proposed MFPT algorithm, we conduct systematic experiments against six representative baseline methods spanning diverse technical paradigms. These include foundational models HRNet48\cite{SunXLW19} and Segformer-B5\cite{DBLP:conf/nips/XieWYAAL21}, state-of-the-art IML algorithms PSCC-Net\cite{DBLP:journals/tcsv/LiuLCL22}, EVP\cite{DBLP:conf/cvpr/LiuSPC23}, and EITLNet\cite{guo2024effective}, and the DIML-specific Patches\cite{Tantaru_2024_WACV} model. All models are trained on the DEAL-300K training set and evaluated on four test subsets: DEAL-E, DEAL-Full, DEAL-A, and DEAL-MB. 

As presented in Table~\ref{tab:table_deal300k}, MFPT demonstrates superior performance across all evaluation scenarios. On the DEAL-E subset for single-round editing scenarios, MFPT achieves 70.30\% IoU and 82.56\% pF1, significantly outperforming all baselines. This validates our multi-frequency prompt tuning mechanism, which effectively captures both high-frequency texture anomalies and semantic inconsistencies to achieve precise pixel-level localization of edited regions. On the DEAL-Full subset containing mixed authentic and tampered samples, MFPT maintains strong performance with 69.86\% IoU and 82.25\% pF1.

For authentic image detection on the DEAL-A subset, MFPT achieves 99.84\% pACC, demonstrating an exceptionally low false positive rate that meets practical forensic requirements. On the challenging DEAL-MB subset featuring multi-round editing scenarios, MFPT achieves 57.68\% pF1, surpassing the second-best method EVP by 17.05\%. This result highlights both the unique value of DEAL-300K in simulating complex social media tampering and the strong generalization capability of vision foundation models enhanced with frequency-domain prompt learning.

\begin{table}[htb]
\centering
\caption{In-Dataset Performance Evaluation}
\label{tab:table_deal300k}
\resizebox{\columnwidth}{!}{%
\begin{tabular}{cccccccc}
\hline
\multirow{2}{*}{Model} & \multicolumn{2}{c}{DEAL-E}      & \multicolumn{2}{c}{DEAL-Full}   & DEAL-A         & \multicolumn{2}{c}{DEAL-MB}     \\ \cline{2-8} 
                        & IoU(\%)        & pF1(\%)        & IoU(\%)        & pF1(\%)        & pACC(\%)       & IoU(\%)        & pF1(\%)        \\ \hline
HRNet48\cite{SunXLW19}                 & 32.90          & 49.51          & 31.99          & 48.47          & 99.36          & 8.53           & 15.73          \\
Segformer-B5\cite{DBLP:conf/nips/XieWYAAL21}            & 39.65          & \underline{56.78}          & 39.63          & \underline{56.76}          & \textbf{99.99}          & 3.07           & 5.96           \\
PSCC-Net\cite{DBLP:journals/tcsv/LiuLCL22}                & 24.58          & 32.87          & 21.07          & 15.53          & 84.71          & \underline{25.95}          & 37.27          \\
EVP\cite{DBLP:conf/cvpr/LiuSPC23}                     & 10.33          & 22.59          & 6.56           & 16.80          & 54.20          & 22.71          & \underline{40.63}          \\
Patches\cite{Tantaru_2024_WACV}                 & 11.53          & 16.58          & 7.54           & 10.85          & 99.12          & 5.48           & 9.02           \\ 
EITLNet\cite{guo2024effective}     & \underline{43.32}          & 55.32          & \underline{40.62}          & 36.20          & 94.53          & 23.81           & 34.50           \\  \hline
MFPT (Ours)                   & \textbf{70.30}          & \textbf{82.56}          & \textbf{69.86}          & \textbf{82.25}          & \underline{99.84}          & \textbf{40.53} & \textbf{57.68} \\ \hline
\end{tabular}%
}
\end{table}

To assess cross-domain generalization capability, we evaluate models trained on DEAL-300K on three public datasets generated by mainstream diffusion models, as shown in Table~\ref{tab:deal2p2}. On the GLIDE-generated CoCoGlide dataset, MFPT achieves 80.97\% pF1, substantially outperforming the second-best EVP method by 36.65\%. For context, TruFor\cite{Tantaru_2024_WACV}, which was specifically designed for CoCoGlide with additional training data, achieves a pF1 of 52.30\%, still significantly lower than MFPT. This confirms both the effectiveness of DEAL-300K and the robustness of our approach.

On the DALL-E2-generated AutoSplice dataset (JPEG100 quality), MFPT achieves 54.76\% pF1, demonstrating competitive performance. For facial attribute editing on Repaint-P2/CelebA-HQ, Segformer-B5 achieves the highest performance with 77.96\% pF1, validating the representativeness of DEAL-300K samples for facial deepfake scenarios. These cross-domain results demonstrate that DEAL-300K, constructed with diverse editing instructions, effectively enhances model adaptability across different diffusion architectures and addresses inter-domain discrepancy challenges that limit traditional forensic methods.

\begin{table}[htb]
\centering
\caption{Cross-Dataset Performance Evaluation}
\label{tab:deal2p2}
\resizebox{\columnwidth}{!}{%
\begin{tabular}{ccccccc}
\hline
\multirow{2}{*}{Model} & \multicolumn{2}{c}{CoCoGlide}   & \multicolumn{2}{c}{AutoSplice (JPEG100)} & \multicolumn{2}{c}{Repaint-P2/CelebA-HQ} \\ \cline{2-7} 
                        & IoU(\%)        & pF1(\%)        & IoU(\%)             & pF1(\%)            & IoU(\%)             & pF1(\%)            \\ \hline
HRNet48\cite{SunXLW19}                 & 22.07          & 36.16          & 35.20               & 52.07              & \underline{59.75}   & \underline{74.80}  \\
Segformer-B5\cite{DBLP:conf/nips/XieWYAAL21}            & 5.25           & 9.98           & 24.54               & 39.40              & \textbf{63.88}      & \textbf{77.96}     \\
PSCC-Net\cite{DBLP:journals/tcsv/LiuLCL22}                & 17.82          & 22.49          & \textbf{42.59}      & 53.75              & 12.29               & 17.25              \\
EVP\cite{DBLP:conf/cvpr/LiuSPC23}                     & 24.72          & \underline{44.32}          & 10.57               & \textbf{55.79}     & 14.18               & 30.23              \\
Patches\cite{Tantaru_2024_WACV}                 & 3.98           & 6.34           & 7.91                & 12.49              & 5.71                & 10.14              \\ 
EITLNet\cite{guo2024effective}      & \underline{33.95}          & 17.99          & 36.85               & 29.12              & 25.17               & 33.63              \\ \hline
MFPT (Ours)                    & \textbf{68.02} & \textbf{80.97} & \underline{37.71}   & \underline{54.76}  & 54.99               & 70.92              \\ \hline
\end{tabular}%
}
\end{table}

\subsection{Robustness Analysis}
To evaluate the robustness of our proposed MFPT algorithm against common image degradations, we conduct comprehensive experiments on two benchmark datasets: DEAL-300K and AutoSplice. We compare against representative baseline methods including Segformer-B5 \cite{DBLP:conf/nips/XieWYAAL21} and EITLNet \cite{guo2024effective}. Our analysis focuses on two prevalent degradation types: JPEG compression (with quality factors ranging from 100 to 50) and Gaussian blur (with kernel sizes from 0 to 19).

\subsubsection{JPEG Compression Robustness}
We first examine model performance under varying JPEG compression levels. Table~\ref{tab:autosplicejpeg} presents results on the AutoSplice dataset. Our MFPT model demonstrates consistent performance across different compression qualities, with moderate compression (JPEG90) yielding a slight performance improvement. This suggests that compression artifacts may provide complementary cues for detecting diffusion-based forgeries when properly modeled.

On the DEAL-300K dataset (Table~\ref{tab:robust_jpeg}), we observe distinct performance patterns across methods. Segformer-B5 shows significant performance degradation as compression intensifies, with IoU decreasing from 39.63\% at JPEG100 to 25.05\% at JPEG50. EITLNet exhibits unstable behavior, showing marginal improvements at medium compression levels (JPEG60-JPEG70) where compression artifacts partially align with editing boundaries. In contrast, MFPT maintains remarkably stable performance across all compression levels, with IoU values remaining above 68.73\% even under severe compression (JPEG50). This stability demonstrates the effectiveness of our multi-frequency prompting mechanism in distinguishing compression artifacts from genuine forensic traces through frequency-domain feature analysis.

\begin{table}[htbp]
\centering
\caption{Localization performance under JPEG compression on AutoSplice dataset}
\label{tab:autosplicejpeg}
\resizebox{0.5\columnwidth}{!}{%
\begin{tabular}{ccc}
\hline
Compression Level & IoU(\%) & F1(\%) \\ \hline
JPEG100 & 37.71 & 54.76 \\
JPEG90  & \textbf{39.21} & \textbf{56.34} \\
JPEG75  & 38.30 & 55.38 \\ \hline
\end{tabular}%
}
\end{table}

\begin{table}[htbp]
\centering
\caption{Robustness against JPEG compression on DEAL-300K dataset}
\label{tab:robust_jpeg}
\resizebox{0.9\columnwidth}{!}{%
\begin{tabular}{c*6{c}}
\hline
\multirow{2}{*}{JPEG Quality} & \multicolumn{2}{c}{Segformer-B5\cite{DBLP:conf/nips/XieWYAAL21}} & \multicolumn{2}{c}{EITLNet\cite{guo2024effective}} & \multicolumn{2}{c}{MFPT (Ours)} \\
\cline{2-3} \cline{4-5} \cline{6-7} 
 & IoU(\%) & pF1(\%) & IoU(\%) & pF1(\%) & IoU(\%) & pF1(\%) \\ \hline
100 & 39.63 & 56.76 & 40.62 & 36.20 & \textbf{69.86} & \textbf{82.25} \\
90  & 40.91 & 58.07 & 35.33 & 30.65 & \textbf{69.84} & \textbf{82.24} \\
80  & 39.88 & 57.02 & 35.17 & 30.41 & \textbf{70.76} & \textbf{82.88} \\
70  & 34.28 & 51.06 & 39.07 & 31.31 & \textbf{69.60} & \textbf{82.08} \\
60  & 27.48 & 43.12 & 42.75 & 31.56 & \textbf{70.32} & \textbf{82.57} \\
50  & 25.05 & 40.07 & 42.57 & 31.40 & \textbf{68.73} & \textbf{81.47} \\ \hline
\end{tabular}%
}
\end{table}

\subsubsection{Gaussian Blur Robustness}
Table~\ref{tab:robust_blur} evaluates model performance under Gaussian blur on the DEAL-300K dataset. As blur intensity increases, all methods experience performance degradation, but MFPT demonstrates superior robustness. At the highest blur level (kernel size 19), MFPT maintains an IoU of 44.60\%, significantly outperforming Segformer-B5 (31.40\%) and EITLNet (13.25\%). EITLNet shows particularly poor robustness, with performance collapsing rapidly as blur intensity increases. MFPT's consistent performance across blur levels confirms the effectiveness of leveraging vision foundation model priors for micro-feature discrimination, highlighting its practical applicability in scenarios involving image blurring.

\begin{table}[htbp]
\centering
\caption{Robustness against Gaussian blur on DEAL-300K dataset}
\label{tab:robust_blur}
\resizebox{0.9\columnwidth}{!}{%
\begin{tabular}{c*6{c}}
\hline
\multirow{2}{*}{Kernel Size} & \multicolumn{2}{c}{Segformer-B5\cite{DBLP:conf/nips/XieWYAAL21}} & \multicolumn{2}{c}{EITLNet\cite{guo2024effective}} & \multicolumn{2}{c}{MFPT (Ours)} \\
\cline{2-3} \cline{4-5} \cline{6-7} 
 & IoU(\%) & F1(\%) & IoU(\%) & F1(\%) & IoU(\%) & F1(\%) \\ \hline
0  & 39.63 & 56.76 & 40.62 & 36.20 & \textbf{69.86} & \textbf{82.25} \\
3  & 43.40 & 60.53 & 33.42 & 28.50 & \textbf{67.84} & \textbf{80.84} \\
7  & 40.22 & 57.37 & 22.38 & 23.32 & \textbf{62.08} & \textbf{76.60} \\
11 & 34.98 & 51.83 & 16.66 & 20.55 & \textbf{56.84} & \textbf{72.48} \\
15 & 34.98 & 51.83 & 14.52 & 18.90 & \textbf{51.08} & \textbf{67.62} \\
19 & 31.40 & 47.80 & 13.25 & 17.66 & \textbf{44.60} & \textbf{61.68} \\ \hline
\end{tabular}%
}
\end{table}

\subsection{Comparison with Pre-trained Models on Existing Benchmarks}

In this section, we evaluated pre-trained localization models using the CoCoGlide and Repaint-P2/CelebA-HQ datasets, comparing our methods with those reported in~\cite{Tantaru_2024_WACV} as shown in Table~\ref{tab:finetunedrp2}. Our model achieved an IoU of $68.0\%$ and pACC of $70.5\%$ on CoCoGlide, indicating superior localization capabilities. In contrast, models~\cite{DBLP:conf/cvpr/0001AN19,DBLP:journals/tifs/CozzolinoV20} trained on traditional datasets for manual editing demonstrated limited IoU performance, emphasizing the need for datasets tailored to diffusion-based image edits. Although Patches~\cite{Tantaru_2024_WACV}, designed for deepfake detection and trained on Repaint-P2/CelebA-HQ, performed well on its training dataset, it struggled to generalize to CoCoGlide. Despite methodological and data source differences from Repaint-P2/CelebA-HQ, our dataset, designed for broader applications and characterized by extensive diversity, maintained exceptional performance. These findings highlight the efficacy and generalization capacity of our algorithm and dataset across various tasks within the DIML domain.

\begin{table}[htbp]
\centering
\caption{Cross-Dataset Performance Evaluation of Pre-trained Models, with other results from~\cite{Tantaru_2024_WACV}. \textcolor{red}{Notably}: Patches~\cite{Tantaru_2024_WACV} is trained on Repaint-P2/CelebA-HQ trainset. }
\label{tab:finetunedrp2}
\resizebox{0.8\columnwidth}{!}{%
\begin{tabular}{ccccc}
\hline
\multirow{2}{*}{Models}                      & \multicolumn{2}{c}{CocoGlide}                   & \multicolumn{2}{c}{Repaint-P2/CelebA-HQ} \\ \cline{2-5} 
                                             & IoU(\%)                & pACC(\%)               & IoU(\%)             & pACC(\%)           \\ \hline
MantraNet~\cite{DBLP:conf/cvpr/0001AN19}     & 25.1                   & 79.8                   & 4.8                 & \textbf{81.9}      \\
Noiseprint~\cite{DBLP:journals/tifs/CozzolinoV20}  & 23.9             & 29.0                   & 18.2                & 23.8               \\
PSCC-Net~\cite{DBLP:journals/tcsv/LiuLCL22}  & 33.3                   & 80.6                   & 14.3                & 66.5               \\
TruFor~\cite{DBLP:conf/cvpr/GuillaroCSDV23}   & 29.2                   & \textbf{81.4}          & 23.1                & 81.3               \\
HiFI-Net~\cite{DBLP:conf/cvpr/GuoLRGM023}    & 2.6                    & 3.2                    & 0.0                 & 81.0               \\
Patches~\cite{Tantaru_2024_WACV}     & 30.8                   & 64.8              & -       & -               \\ \hline
Ours                                         & \textbf{68.0}          & 70.5                   & \textbf{54.9}       & 78.3               \\ \hline
\end{tabular}%
}
\end{table}

\subsection{Evaluation of Pre-trained Models on DEAL-300K}
To assess the challenging nature of the DEAL-300K dataset, we evaluate four state-of-the-art image manipulation localization algorithms using their publicly available pre-trained models without fine-tuning. The results are presented in Table~\ref{tab:pretrainedmodel}. We note that direct performance comparison using pre-trained models from different datasets may lack fairness due to domain mismatches. This experiment primarily aims to highlight the distinctive characteristics of DEAL-300K compared to existing benchmarks.

PSCC-Net\cite{DBLP:journals/tcsv/LiuLCL22}, EVP\cite{DBLP:conf/cvpr/LiuSPC23}, and EITLNet\cite{guo2024effective}, trained on manually edited datasets, along with Patches\cite{Tantaru_2024_WACV}, developed specifically for diffusion-based facial editing, all exhibit limited performance on DEAL-300K. Their poor performance stems primarily from their inability to effectively handle diffusion-edited images that exhibit more realistic visual effects and diverse editing scenarios. Notably, EVP achieves the highest pF1 scores on DEAL-E (22.59\%) and DEAL-Full (16.45\%) among the baselines, demonstrating some adaptability to diffusion artifacts through its frequency-domain processing. However, EVP shows significant weakness on authentic images (DEAL-A), achieving only 36.86\% pACC due to its base network's inability to accurately distinguish complex semantic features, resulting in high false positive rates. This observation reveals that while frequency-domain approaches provide some robustness to diffusion artifacts, more sophisticated semantic understanding is required for accurate localization in realistic editing scenarios.

\begin{table}[htbp]
\centering
\caption{Evaluation of pre-trained localization models on DEAL-300K.}
\label{tab:pretrainedmodel}
\resizebox{\columnwidth}{!}{%
\begin{tabular}{cccccccc}
\hline
\multirow{2}{*}{Model} & \multicolumn{2}{c}{DEAL-E} & \multicolumn{2}{c}{DEAL-Full} & DEAL-A & \multicolumn{2}{c}{DEAL-MB} \\ \cline{2-8}
& IoU(\%) & pF1(\%) & IoU(\%) & pF1(\%) & pACC(\%) & IoU(\%) & pF1(\%) \\ \hline 
PSCC-Net\cite{DBLP:journals/tcsv/LiuLCL22} & 1.31 & 2.47 & 1.02 & 2.03 & \textbf{99.90} & 19.09 & 28.29 \\
EVP\cite{DBLP:conf/cvpr/LiuSPC23} & \textbf{8.65} & \textbf{22.59} & 5.88 & \textbf{16.45} & 36.86 & 20.43 & 39.10 \\
Patches\cite{Tantaru_2024_WACV} & 1.56 & 2.02 & 1.50 & 1.94 & 99.22 & 22.55 & 33.30 \\ 
EITLNet\cite{guo2024effective} & 7.19 & 9.52 & \textbf{21.78} & 6.23 & 97.91 & \textbf{36.45} & \textbf{49.79} \\
\hline
\end{tabular}%
}
\end{table}

\subsection{Ablation Study}

\begin{table*}[htbp]
\centering
\caption{Performance Comparison of Various Pre-trained VFMs within Our Model.}
\label{tab:vfms}
\resizebox{1.5\columnwidth}{!}{%
\begin{tabular}{ccccccccccccc}
\hline
\multirow{2}{*}{Pre-trained Models} & \multicolumn{2}{c}{DEAL-E}      & \multicolumn{2}{c}{DEAL-Full}   & \multicolumn{2}{c}{DEAL-MB}     & \multicolumn{2}{c}{CoCoGlide}   & \multicolumn{2}{c}{AutoSplice (JPEG100)} & \multicolumn{2}{c}{R-P2/CelebA} \\ \cline{2-13} 
                        & IoU(\%)        & pF1(\%)        & IoU(\%)        & pF1(\%)        & IoU(\%)        & pF1(\%)        & IoU(\%)        & pF1(\%)        & IoU(\%)             & pF1(\%)            & IoU(\%)        & pF1(\%)        \\ \hline
EVA02~\cite{DBLP:journals/corr/abs-2303-11331}                  & 67.35          & 80.49          & 65.86          & 79.42          & 38.74          & 55.85          & 65.94          & 79.48          & \textbf{53.24}      & \textbf{69.48}     &53.87                & 70.02               \\
DINOv2~\cite{DBLP:journals/corr/abs-2304-07193}                  & \textbf{70.30} & \textbf{82.56} & \textbf{69.86} & \textbf{82.25} & \textbf{40.53} & \textbf{57.68} & \textbf{68.02} & \textbf{80.97} & 37.71               & 54.76              & \textbf{54.99}          & \textbf{70.92}          \\ \hline
\end{tabular}%
}
\end{table*}

In this section, we provide detailed analysis on the effectiveness of our core model designs. 
Table~\ref{tab:vfms} compares the performance of EVA02~\cite{DBLP:journals/corr/abs-2303-11331} and DINOv2~\cite{DBLP:journals/corr/abs-2304-07193} as frozen encoders in various datasets. Though both are based on the ViT model with differing pre-training strategies, DINOv2 demonstrates superior performance across most datasets, notably achieving the highest IoU and pF1 scores on the DEAL-Full dataset. In contrast, EVA02 shows competitive performance on the AutoSplice dataset, highlighting the importance of selecting the appropriate VFM to improve the accuracy and efficiency of DIML algorithms in different scenarios. Additionally, with a total parameter count of 331.86M, of which 27.62M are trainable, representing a ratio of $8.3\%$, this emphasizes the significant role of the prior knowledge embedded in VFMs for DIML tasks.

\begin{table}[htbp]
\centering
\caption{Ablation Study on Different Components. Baseline is Mask2Former with Frozen DINOv2.}
\label{tab:Ablation}
\resizebox{\columnwidth}{!}{%
\begin{tabular}{cccccllcc}
\hline
\multirow{2}{*}{Models} & \multicolumn{2}{c}{DEAL-E} & \multicolumn{2}{c}{DEAL-Full} & \multicolumn{2}{c}{DEAL-MB}                               & \multicolumn{2}{c}{CoCoGlide} \\ \cline{2-9} 
                        & IoU(\%)      & pF1(\%)     & IoU(\%)       & pF1(\%)       & \multicolumn{1}{c}{IoU(\%)} & \multicolumn{1}{c}{pF1(\%)} & IoU(\%)       & pF1(\%)       \\ \hline
Baseline~(Frozen Encoder)~\cite{DBLP:conf/cvpr/ChengMSKG22}   & 3.28       & 6.35       & 2.86         & 5.57         & 23.82                       & 38.48                       & 3.12         & 6.04         \\
Baseline+Rein Adapter~\cite{DBLP:journals/corr/abs-2312-04265}   & 70.88        & 82.96       & 70.53         & 82.72         & 33.82                       & 50.54                       & 63.69         & 77.82         \\
Baseline+Rein Adapter+FFrP           & 65.36        & 79.05       & 64.24         & 78.23         & 42.37                       & 59.52                       & 70.38         & 82.62         \\
Baseline+Rein Adapter+ FInP + FFrP   & 70.30        & 82.56       & 69.86         & 82.25         & 40.53                       & 57.68                       & 68.02         & 80.97         \\ \hline
\end{tabular}%
}
\end{table}

\begin{figure}[htb]
\centering 
\includegraphics[width=\columnwidth]{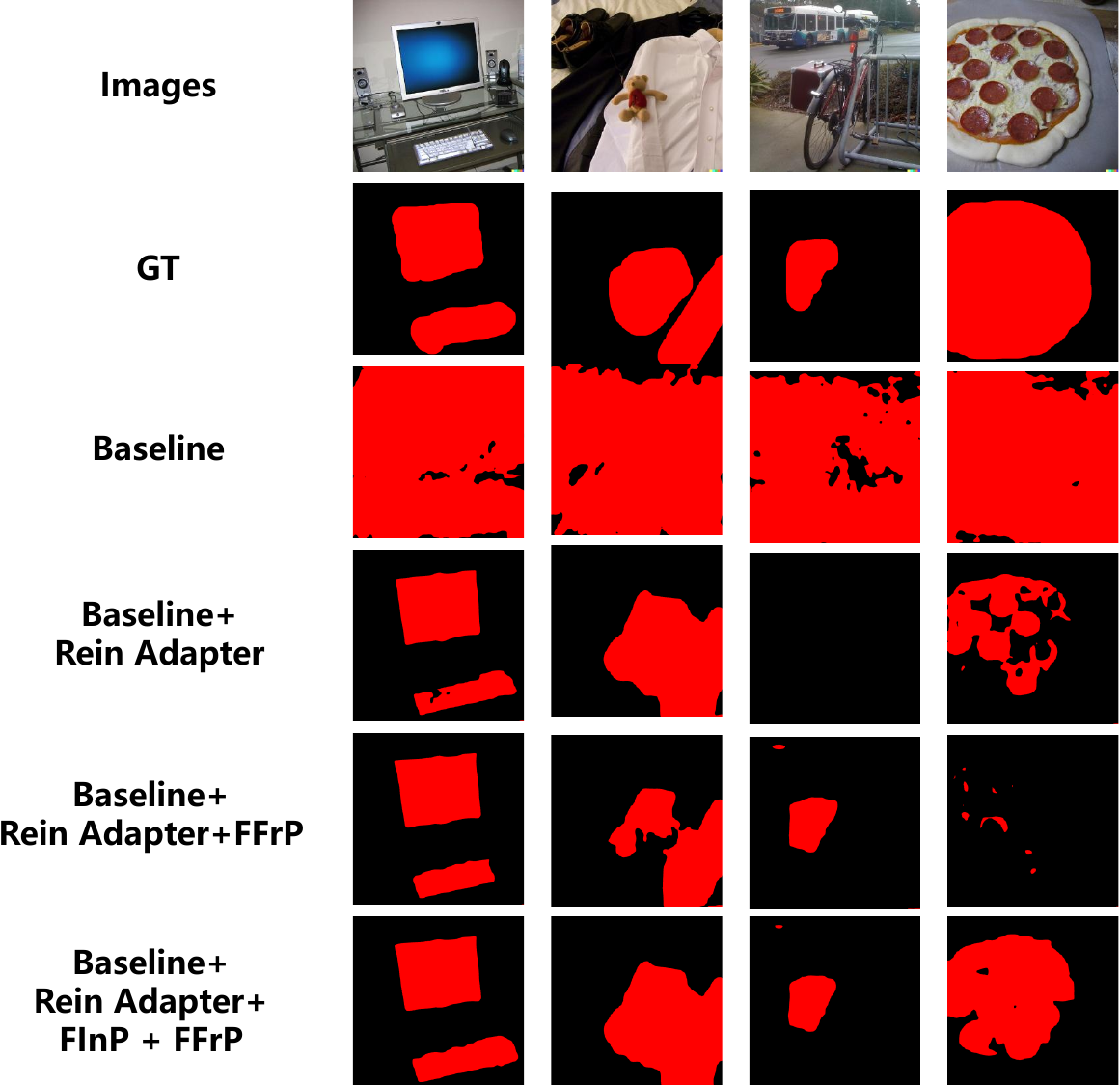} 
\caption{Visualization of Ablation Study Results on DEAL-MB Dataset}
\label{dealmg_abl_result}
\end{figure}

Table~\ref{tab:Ablation} presents an ablation study examining the influence of various components on the performance of our model. The baseline model utilizes Mask2Former with a frozen DINOv2 as the encoder. While the baseline with Rein Adapter demonstrates proficiency in in-domain tests, it exhibits limitations in out-of-domain scenarios. The addition of FFrP notably enhances out-of-domain performance, as evidenced by the improved metrics on CoCoGlide and DEAL-MB. Our final model, incorporating both FInP and FFrP, achieves a balanced and superior performance across all datasets. This underscores the capability of our MFPT architecture to deliver competitive results in diverse settings, showcasing its adaptability and effectiveness in DIML tasks.Figure~\ref{dealmg_abl_result} further presents the qualitative results of different modules on the DEAL-MB dataset. It is evident that the final model achieves superior localization of manipulated regions generated by diffusion models, demonstrating effective pixel-level forgery detection capability.

\subsection{Visualization of Localization Results}
\begin{figure}[htbp]
\centering
\includegraphics[width=\columnwidth]{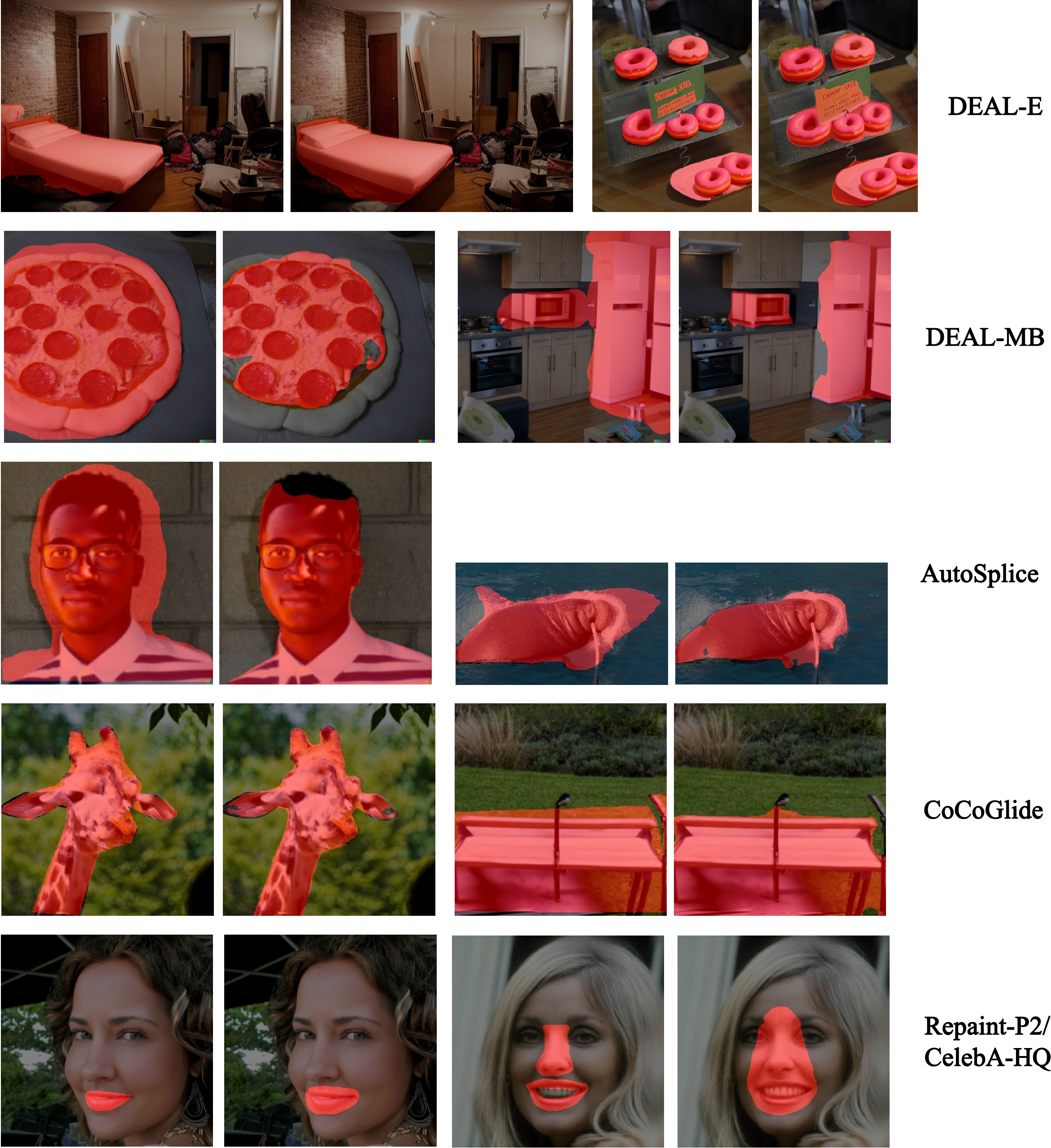}
\caption{Qualitative results of our model on all datasets. For every pair of images, the left is the ground-truth and the right is our prediction.}
\label{moresamples}
\end{figure}

 We present the qualitative localization results of our model in various reference datasets in Figure~\ref{moresamples}. Our approach excels at identifying specific tampered regions, surpassing mere saliency or anomaly detection. It accurately delineates the extent of regions created through diffusion. However, the current lack of detail refinement hampers optimal performance. Further optimization efforts are planned to address this in future work.

\section{Conclusion}
This paper introduced DEAL-300K, the largest dataset to date for localizing diffusion-based image manipulations, alongside a novel annotation process that facilitates future dataset expansion. We proposed a framework utilizing Visual Foundation Models (VFMs) with Multi-Frequency Prompt Tuning (MFPT) to detect subtle abnormalities in edited areas through low-level information. Comprehensive experiments across DEAL-300K and existing benchmark datasets validated our approach, demonstrating strong in-domain performance, cross-domain generalization, and robustness against JPEG compression. These results underscore the effectiveness of our model and the value of our dataset. Moreover, our automated annotation method, which will be open-sourced, promises to significantly reduce the cost of annotating larger datasets, supporting future advancements in this field. In the fearture, we plan to extend DEAL-300K to include video manipulations, further enhancing its applicability to real-world scenarios.

\bibliographystyle{IEEEtran}
\bibliography{ref.bib}

\end{document}